%% file: main.tex
\documentclass{article}

\usepackage{microtype}
\usepackage{graphicx}
\usepackage{subcaption}
\usepackage{booktabs} 

\usepackage{enumitem} 
\usepackage{xspace}
\usepackage{bbm}
\usepackage{dsfont}
\usepackage{verbatim}
\usepackage{caption}

\usepackage[T1]{fontenc}

\usepackage[hashEnumerators,smartEllipses]{markdown} 

\usepackage{xurl}

\usepackage{hyperref}

\usepackage[accepted]{icml2024}

\usepackage{amsthm}

\newtheoremstyle{defstyle}
  {0.1cm}
  {0.1cm}
  {}
  {}
  {}
  {:}
  {.5em}
  {\thmnumber{(#2) }\thmname{\textit{#1}}\thmnote{ \textit{#3}}}

\theoremstyle{defstyle}
\newtheorem{definition}{}

\icmltitlerunning{Position: A Roadmap to Pluralistic Alignment}

\usepackage{etoolbox}

\begin{document}

\twocolumn[
\icmltitle{Position: A Roadmap to Pluralistic Alignment}
\icmlsetsymbol{equal}{*}

\begin{icmlauthorlist}
\icmlauthor{Taylor Sorensen}{uw}
\icmlauthor{Jared Moore}{stanford}
\icmlauthor{Jillian Fisher}{uw,uwstats}
\icmlauthor{Mitchell Gordon}{uw,mit}
\icmlauthor{Niloofar Mireshghallah}{uw}
\icmlauthor{Christopher Michael Rytting}{uw}
\icmlauthor{Andre Ye}{uw}
\icmlauthor{Liwei Jiang}{uw,ai2}
\icmlauthor{Ximing Lu}{uw}
\icmlauthor{Nouha Dziri}{ai2}
\icmlauthor{Tim Althoff}{uw}
\icmlauthor{Yejin Choi}{uw,ai2}
\end{icmlauthorlist}

\icmlaffiliation{uw}{Department of Computer Science, University of Washington, Seattle, Washington, USA}
\icmlaffiliation{uwstats}{Department of Statistics, University of Washington, Seattle, Washington, USA}
\icmlaffiliation{stanford}{Department of Computer Science, Stanford University, Stanford, California, USA}
\icmlaffiliation{mit}{Department of Electrical Engineering and Computer Science, MIT, Cambridge, Massachusetts, USA}
\icmlaffiliation{ai2}{Allen Institute for Artificial Intelligence, Seattle, Washington, USA}

\icmlcorrespondingauthor{Taylor Sorensen}{tsor13@cs.washington.edu}
\icmlcorrespondingauthor{Yejin Choi}{yejin@cs.washington.edu}

\icmlkeywords{Machine Learning, ICML, pluralism, value pluralism, alignment, llm, nlp, rlhf, ethics, fairness, accountability}

\vskip 0.3in
]



\printAffiliationsAndNotice{}  


\begin{abstract}
With increased power and prevalence of AI systems, it is ever more critical that AI systems are designed to serve \emph{all}, i.e., people with diverse values and perspectives.
However, aligning models to serve \textit{pluralistic} human values
remains an 
open research question.
In this piece, we propose a roadmap to pluralistic alignment, specifically using large language models as a test bed.
We
identify and formalize  
three possible ways to define and operationalize pluralism in AI systems:
1) \textit{Overton pluralistic} models that present a spectrum of reasonable responses;
2) \textit{Steerably pluralistic} models that can steer to reflect certain perspectives; and
3) \textit{Distributionally pluralistic} models that are well-calibrated to a given population in distribution.
We also formalize and discuss three possible classes of \textit{pluralistic benchmarks}:
1) \textit{Multi-objective} benchmarks,
2) \textit{Trade-off steerable} benchmarks that incentivize models to steer to arbitrary trade-offs,
and 3) \textit{Jury-pluralistic} benchmarks that explicitly model diverse human ratings.
We use this framework to argue that current alignment techniques may be fundamentally limited for pluralistic AI; indeed, we highlight empirical evidence, both from our own experiments and from other work, that standard alignment procedures might \textit{reduce} distributional pluralism in models,
motivating the need for further research on pluralistic alignment.

\end{abstract}

\begin{figure}[t]
    \centering
    \includegraphics[width=1\linewidth]{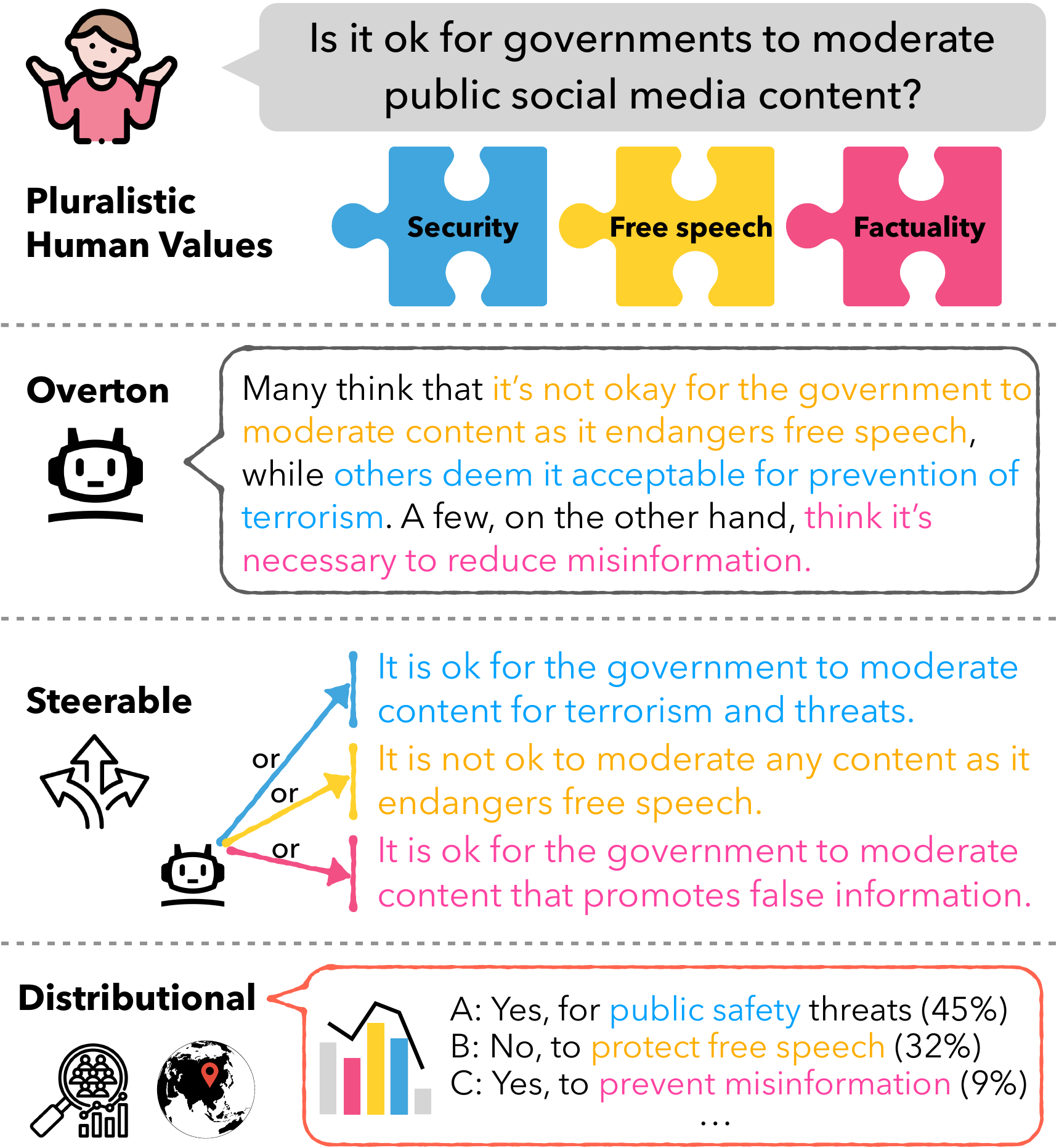}
    \caption{Three kinds of pluralism in models.}
    \label{fig:definitions}
\end{figure}

\section{Introduction}
AI alignment aims to ensure that a system works with human intentions and values \cite{leike2018scalable, ji2024ai, gabriel2020}.  However, even within a single task or prompt, individual people vary widely in their goals, intentions, and values.  As a broader set of people use and rely upon AI systems, we need systems that can understand and cater to a broader set of needs.  In other words, we need systems that are \textit{pluralistic}, or capable of representing a diverse set of human values and perspectives. While many in the community have argued for this \cite{bai2022constitutional,Gordon_2022,sorensen2023value}, at least two important questions remain: \textit{How, concretely, can a system be pluralistic?} and \textit{How might benchmarks be designed to measure pluralism}?

\textbf{In this piece, we advocate for explicit pluralistic considerations in aligning AI systems} (\S \ref{sec:pluralism}).
In particular, we use large language models (LLMs) as a testbed for alignment \cite{askell2021general}, though we believe the concepts can generalize to other AI systems (\S \ref{subsec:generalize}). Because pluralism may look different in different contexts, we formalize three distinct ways of operationalizing pluralism for AI systems/models: 1) providing comprehensive, high-coverage responses (Overton pluralism, \S \ref{sec:overton-pluralism}), 2) an ability to be faithfully steered to represent particular attributes (steerable pluralism, \S \ref{sec:steerable-pluralism}), and 3) distributional representation of a population (distributional pluralism, \S \ref{sec:distributional-pluralism}). Each form of pluralism has cases where they may be desirable to maximize. We also define three types of pluralistic benchmarks: multi-objective benchmarks (\S \ref{sec:multi-objective-benchmarks}), benchmarks of models' steerability across objectives (trade-off steerable benchmarks, \S \ref{sec:trade-off-steerable-benchmarks}), and benchmarks that explicitly model individuals (jury-pluralistic benchmarks, \S \ref{sec:jury-pluralistic-benchmarks}). We also outline the situations for which each would be useful.

We then discuss the relationship between current alignment approaches and pluralism (\S \ref{sec:current-alignment}) and provide initial findings that current alignment techniques \textit{reduce}  distributional  pluralism. We advocate and lay out a plan for future work toward pluralistic evaluations and alignment.

\section{Arguments for Pluralism in AI Systems}
\label{sec:pluralism}
In this section, we argue for the importance of pluralism in aligning AI models.

\textbf{Customization necessitates pluralism.} Any guardrails placed on AI systems will require  customization, within the bounds of those guardrails, to serve diverse use cases and values \cite{chen2023large, jang2023personalized}.  Pluralism can illuminate the set of values or attributes that users may customize to, and provide an understanding of how well a system can be steered (\S \ref{sec:steerable-pluralism}, \ref{sec:trade-off-steerable-benchmarks}).

\textbf{Pluralistic systems have technical benefits.} Implicit to current preference-based methods like reinforcement learning with human feedback (RLHF) is the assumption that models should fit to the ``average'' human preference. However, this treats human variation as noise instead of signal \cite{aroyo2023dices, siththaranjan2023distributional} -- pluralism, however, recognizes this as signal. Modeling pluralism also may increase interpretability by enabling a clearer relationship between decisions and their source (\S \ref{sec:steerable-pluralism}, \ref{sec:trade-off-steerable-benchmarks}).

\textbf{Pluralistic evaluations enable generalist systems.} Recently, AI/NLP has trended away from specialist systems and towards generalist systems (foundation models) for use in a diverse set of tasks by a diverse set of users. Yet, current alignment optimizes these generalist systems for a single objective -- averaged human preferences.  To understand the strengths and weaknesses of these systems, we must measure how they perform across a variety of objectives (\S \ref{sec:multi-objective-benchmarks}) \cite{ethayarajh-jurafsky-2022-authenticity} and users (\S \ref{sec:steerable-pluralism}, \ref{sec:distributional-pluralism}, \ref{sec:jury-pluralistic-benchmarks}).

\textbf{Pluralism as a value itself.}
Many modern societies view accepting competing values and perspectives as a core value in and of itself. Theorists have extolled the benefits of political pluralism \cite{tocqueville1835democracy,berlin1969two,rawls1996political}, moral and value pluralism \cite{nagel1979fragmentation,kekes1993morality,raz1999engaging}, and pluralist theories of truth \cite{wright1992truth,sher1998possibility}. While this piece primarily focuses on surfacing differing ideas, perspectives, and values (\S \ref{sec:models}, \ref{sec:benchmarks}), our scaffolding for technical measurements and implementations of value can also apply to other notions of pluralism. This stands in contrast to current alignment procedures such as RLHF which have been characterized as implementing ``preference-based utilitarianism.'' \cite{tasioulas}.

\textbf{AI systems should reflect human diversity.}
We contend that AI systems should reflect and support the diversity amongst humans and their values, as it is both a feature and a desired quality of human societies (\S \ref{sec:distributional-pluralism}, \ref{sec:jury-pluralistic-benchmarks}). Exposure to diverse ideas (\S \ref{sec:overton-pluralism}) also improves deliberation \cite{bowman2022measuring, landemore2015deliberation}. Furthermore, algorithmic monocultures lead to increased unfairness when applied by many decision makers \cite{Bommasani2021OnTO}.

\section{Pluralism for AI Models/Systems}

\label{sec:models}
In this section, we formalize three definitions for how a single model or system can be pluralistic. Specifically, we outline \textit{Overton pluralism}, wherein a model outputs the whole spectrum of reasonable responses; \textit{Steerable pluralism}, wherein a model is faithfully steered to reflect certain properties or perspectives; and \textit{Distributional pluralism}, wherein a model's distribution over answers matches that of a given target population (see Figure \ref{fig:definitions}). For each, we will also discuss relevant applications and potential evaluations, along with limitations and recommendations for future research. 

Throughout, we will consider a model or system $\mathcal{M}$, a query $x$, and a response $y$.  While we specifically focus on natural language queries and responses with $\mathcal{M}$ being an LLM, our definitions can nevertheless generalize to other  inputs, outputs, and models as well.

\subsection{Overton Pluralistic Models}
\label{sec:overton-pluralism}

Given an input, there are often many potential types (or modes) of answers a model can produce. For example, if a user poses a query to an LLM for which there is no single established \textit{correct} answer, the LLM may answer with any one of several \textit{reasonable} answers.

\textbf{Definitions}
Given a query $x$, consider possible answers $y$. 

\begin{definition}[Correct Answer in $\mathcal{C}$]
\label{def:correct}
An answer which can be conclusively verified or with which the overwhelming majority of people across various backgrounds would agree. 
\end{definition}

\begin{definition}[Reasonable Answer in $\mathcal{R}$]
\label{def:reasonable}
An answer for which there is suggestive, but inconclusive, evidence, or one with which significant swaths of the population would agree.
Additional top-down restrictions (e.g., safety) may apply. 
\end{definition}

\begin{definition}[Overton window]
    The set of all reasonable answers: $W(x) = \{y \in \mathcal{Y} \vert (x,y) \in \mathcal{R} \}$.\footnote{Our terminology generalizes the concept of an ``Overton window" as used in political science: ``the spectrum of ideas on public policy and social issues considered acceptable or viable by the general public at a given time." \cite{OED-overton-window}}
\end{definition}

\begin{definition}[A response set $\{y\}$ to a query $x$ is Overton-pluralistic]
    $\{y\}$ contains all potentially reasonable answers in the Overton window. This is in contrast to picking just one answer in the Overton window, or presenting an unreasonable answer which would lie outside the Overton window. A single response may be Overton-pluralistic if it synthesizes the whole response set $\{y\}$.
\end{definition}

\begin{definition}[Model $\mathcal{M}$ is Overton-pluralistic]
    $\mathcal{M}$ gives \textit{Overton-pluralistic} responses to queries, that is for a given input $x$, the output of $\mathcal{M}(x) = W(x)$.
\end{definition}

\textbf{Motivation} 
In many situations, there are many reasonable answers to a question \cite{min-etal-2020-ambigqa,scherrer2023evaluating}. Rather than outputting a single reasonable answer, which may be selected idiosyncratically or in a biased fashion, Overton-pluralistic models output all reasonable answers.

\textbf{Potential Implementation}
We outline two ways to operationalize \textit{Overton pluralism}.
In order to determine an Overton window for a set of queries $X$, one could survey a population for responses to a question and identify clusters (e.g., using semantic similarity) of candidate reasonable answers. Then, one could narrow down the window to reasonable answers $W(x)$ with additional polling for reasonableness, defining a minimum threshold of support, or some other top-down way of filtering out unreasonable responses.
One could define a way to extract the set of ``answers" $\{y\}$ from a model response and compare it to the window.
Alternatively, one could enumerate a list of unreasonable answers $U(x)$ and detect which reasonable or unreasonable answers the response entails 
with an entailment model \cite{Shajalal_2023, liu2022wanli}.
With both methods, metrics like precision/recall/accuracy can be calculated.

\textbf{Applications}
Many relevant domains fall under advice-giving.  Current LLMs often give advice confidently but inconsistently or in an opinionated manner, affecting users' downstream judgments \cite{inconsistent-moral,10.1145/3544548.3581196}. Overton-pluralism requires consideration of multiple heterogeneous judgements, encouraging deliberation over spontaneous judgement \citep{critiquepracticalreason, theoryofjustice}.  It could also aid in scalable oversight \citep{bowman2022measuring} to help users annotate model outputs, in the single ground truth case \cite{michael2023debate} or when we want a diversity of views. Further examples include settings where we want to encourage multiple approaches, such as mathematical proof writing.

\textbf{Limitations}
Defining and operationalizing the Overton window may present a challenge. If a reasonable answer is determined by a set of expert annotators, it may be difficult to scale. If the Overton window is not properly defined, models may contribute to bothsidesism / false balance  \cite{falseBalance,BOYKOFF2004125}. One remedy may be to present the support or certainty for each reasonable answer in addition to its content, although current LLMs struggle with this \cite{zhou2024relying}. Also, while pluralism may never be completely neutral, it can be considered a fairer response to queries \citep{situatedKnowledges}.  Finally, this framework requires long-form responses with multiple answers; other concepts of pluralism may be required for distributions over short answers (see \S \ref{sec:distributional-pluralism}).

\textbf{Alignment Procedures and Recommendations}
While RLHF may \textit{implicitly} steer models to Overton pluralism to the extent that users prefer it, further study into this is needed. Alternatively, one approach to \textit{explicitly} encourage Overton pluralism is taking multiple samples from a model \cite{long2023large, jung2022maieutic}, potentially prompting for diverse outputs \cite{Hayati2023HowFC}, to simulate an Overton window. Alternatively, one could manually create the batch of reasonable responses.  A model can be trained to output a synthesis of the entire batch. Datasets which identify human values \citep{hendrycks2020aligning, sorensen2023value} can be used to evaluate Overton-pluralism. We recommend further study into models' current degree of Overton-pluralism and how it can be amplified for relevant applications.

\subsection{Steerable Pluralistic Models}
\label{sec:steerable-pluralism}

A pluralistic model might instead faithfully \textit{steer} (or align) its responses to a given attribute or perspective, such as a value, framework, or population.

\textbf{Definitions} With this in mind, let us consider:

\begin{definition}[Steering attributes $A$]
    Attributes/properties/perspecti-ves which we wish a model to faithfully reflect. Examples include groups of people from a shared culture, philosophical/political schools of thought, or particular values. To reflect multiple attributes simultaneously, the elements of $A$ could be construed as \textit{sets} of attributes.
\end{definition}

\begin{definition}[Response $y_{\vert x, a}$ faithfully reflects attribute $a\in A$]
    The response $y$ to the query $x$ is consistent with, or follows from, attribute $a$.
\end{definition}

\begin{definition}[Model $\mathcal{M}$ is steerably-pluralistic with respect to attributes $A$]
    Given an input $x$ and an attribute $a \in A$, the model $\mathcal{M}(x, a)$ conditioned on $a$ produces a response $y$ which faithfully reflects $a$.
\end{definition}

\textbf{Motivation}
In many instances, we want models to respond to queries in a consistent and specifiable manner.
Models which have been so heavily ``aligned'' towards a specific attribute such that they cannot be steered to other attributes fail to be useful (or usable) to populations who may not share that value or attribute. We see evidence of this in the ``Silicon Valley'' and ``WEIRD'' \cite{weird} bias of many LLMs, which often skew male, White, American, liberal, and wealthy in perspective \cite{santurkar2023opinions,hartmann2023political,perez2022discovering,santy2023nlpositionality}.

\textbf{Potential Implementation}
Given queries $X$ and attributes $A$, one needs a way to condition the model on attributes at inference. To measure whether a response reflects $a$, one could  either use direct human annotations or reward models that are tuned specifically to the attributes, such as a value-specific reward \cite{sorensen2023value}. These attribute-specific faithfulness scores would be the degree to which a model is steerably pluralistic.

Different attributes may require different metrics for faithfulness, depending on the kind of attribute and level of ambiguity. For example, for a particular difficult moral quandary, there may be no ambiguity given a particular ethical framework (e.g., only one ``correct" or faithful answer). However, if you condition instead on a population, there may still be disagreement or ambiguity - other approaches like an Overton window may apply here.

Several previous works have measured forms of steerable pluralism, particularly with respect to moral, political, and cultural perspectives \citep{out-of-one, jiang2022communitylm, simmons-2023-moral, ramezani2023knowledge, santy2023nlpositionality}. However, previous work suggests that conditional pluralism is far from solved \cite{santurkar2023opinions}.

\textbf{Applications}
An important application of steerable-pluralism is customization. Users often want to personalize models towards characteristic properties and perspectives \cite{chen2023large}, in tasks such as writing assistance \cite{li2023teach} and ideation \cite{Girotra2023,ma2023chatbots}. Steering towards therapeutic values can help in the mental health domain \cite{song2024typing,sharma-rephrase}. Steering models to represent multiple different perspectives can be valuable in creative production \citep{shanahan2023evaluating}, psychological inquiry \citep{shanahan2023roleplay}, simulating social systems \citep{park2022social}, and deliberative discourse \citep{dontJustTellMe, landemore2015deliberation, page2019diversity, page2008difference}.

Moreover, steerably pluralistic models may have useful representations in a variety of settings, such as hate speech detection \cite{feng2023pretraining} and negative thought reframing \cite{sharma-etal-2023-cognitive, Sharma2023FacilitatingSM}. In general, this may allow varying ``cognitive architectures'' for more structured and generally intelligent systems \citep{sumers2023cognitive}.

\textbf{Limitations}
Steerable pluralism requires deciding which attributes are acceptable to steer the model. We may want to disallow some attributes (e.g., hate speech). The challenges here are similar to those in determining which answers are ``reasonable'' in Overton-pluralism, such as subjectivity or arbitrariness in the selection of steerable attributes. Moreover, if attributes are defined too broadly, there is a risk of stereotyping or ``flattening'' the nuances of the complex perspectives and people that attributes are intended to represent \citep{durmus2023measuring}. In some cases, an intersectional evaluation \cite{crenshaw1989demarginalizing}, in which attributes are not considered independently but in conjunction with each other, may be necessary.

\textbf{Alignment Procedures and Recommendations}
There are a variety of ways to induce particular values at inference time. These include conditioning on certain groups \citep{out-of-one, hwang_aligning_2023} and studying which conditions (responses, demographics, etc.) yield the best agreement. \citet{li_steerability_2023, kim2023ai} learn user embeddings which they use to induce certain values from LLMs. \citet{zhao_group_2023} add a module to base LLMs which aims to predict group responses in a few-shot manner. \citet{fleisig_when_2023} predict annotator ratings for specific groups. \citet{Sharma2023FacilitatingSM,sharma2023cognitive} rewrite responses for specific audiences.

We believe that steerability research will become increasingly important as users desire more customizability. While there may be certain behaviors to which a model should not be aligned, we advocate for systems that can be aligned to many attributes within an acceptable range. 

\subsection{Distributionally Pluralistic Models}
\label{sec:distributional-pluralism}

Another way to operationalize pluralism is in the \textit{distribution} over answers compared to a given population.

\textbf{Definitions} In this framework, we consider:

\begin{definition}[A population or group of people $G$]
    A set of people which we want the model to represent.
\end{definition}

\begin{definition}[Model $\mathcal{M}$ is distributionally-pluralistic with respect to a reference population $G$]
    For a given prompt $x$, $\mathcal{M}$ is as likely to provide response $y$ as the reference population $G$. In other words, $\mathcal{M}$ is well-calibrated w.r.t. the distribution over answers from $G$.
\end{definition}

\textbf{Motivation and Applications} 
Distributional pluralism in an LLM is crucial for any application where $\mathcal{M}$ is used to simulate, interface with, or otherwise model the views of a population, e.g., simulating populations via  agent-based modeling \cite{tornberg2023simulating, park2022social, park2023generative}, piloting subject/user responses to surveys \cite{out-of-one, aher2023using}, survey design \cite{ziems2023can}, or studying the internet as a cultural artifact \cite{Buttrick2024}. 

\textbf{Potential Implementation}
Let $X$ be a set of queries to which $G$ gives a distribution $Y$. For example, a census survey or public opinion poll. $\mathcal{M}$'s estimate, $\hat{Y}$, can be compared to the population distribution using any distributional divergence metrics, such as Jensen-Shannon divergence, KL-divergence, or Wasserstein distance \cite{santurkar2023opinions, durmus2023measuring}, or hard measures like accuracy or tetrachoric correlation \cite{out-of-one}.

\textbf{Limitations}
One potential limitation of distributional pluralism is its proportional nature. This means that more frequent opinions will be output by a model with higher frequency, even if this response is harmful - although might be mitigated by defining a window of reasonableness as in Overton pluralism. Another limitation is the need for a predetermined target distribution--a population. In creation of a general LLM, like ChatGPT, who is the target distribution? Furthermore, for many open-ended queries, it is not clear whether there is any response frequency data.

\textbf{Alignment procedures}
While, to our knowledge, there are no alignment procedures to explicitly increase distributional calibration, there are a couple promising directions. One is to simply (pre)train a model on more data from the target population. As the cross entropy objective encourages a model to learn the distributions of speech of a training population, simply providing more data from that population ought to lead to better representation. Another promising direction is to train on the data from a population (e.g., survey data) that one could use to evaluate distributional pluralism, although it is unclear how well this will generalize to novel questions/domains. Further research is needed here.

\textbf{Recommendations}
Oftentimes when researchers measure to which group of people a model best aligns, they compare average responses. In contrast, we advocate for comparing \textit{distributions} because it leads to clearer results: groups of people have distributions over answers, and probabilistic models do as well. We advocate for more distributionally pluralistic evaluations with respect to clearly specified groups of people to better characterize current models. Nonetheless, the stochasticity in distributional pluralism is not desirable in all cases--for example, when the behavior of a model needs to be tightly controlled.

\section{Pluralism for Benchmarks}
\label{sec:benchmarks}

\begin{figure}[t]
    \centering
    \includegraphics[width=1\linewidth]{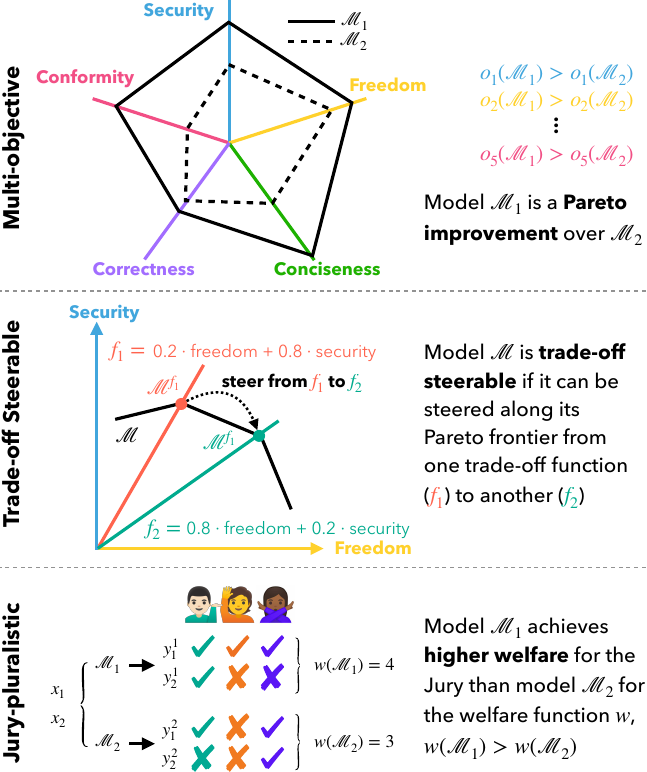}
    \caption{Three kinds of pluralistic benchmarks.}
    \label{fig:benchmarks}
\end{figure}

While the last section defined how a \textit{model} can be pluralistic, here we explore how a \textit{benchmark} can be pluralistic. Most current benchmarks are \textit{monistic} (focused on a single objective). \textit{Pluralistic} benchmarks have \textit{more than one} objective to maximize. Importantly, each is measured separately.

\subsection{Multi-Objective Benchmarks}
\label{sec:multi-objective-benchmarks}

\textbf{Definitions} Define:

\begin{definition}[Objectives to maximize $O=\{o_1, \hdots, o_n\}$]
    A set of multiple objectives to evaluate a model $\mathcal{M}$, each of which which we desire to maximize. Each $o$ maps from a model $\mathcal{M}$ to a scalar in $\mathbb{R}$.
\end{definition}

\begin{definition}[Model $\mathcal{M}_1$ is a Pareto improvement to model $\mathcal{M}_2$.]
    $\forall o_i \in O, o_i(\mathcal{M}_1) \geq o_i(\mathcal{M}_2); \exists o_j \textrm{ s.t. } o_j(\mathcal{M}_1) > o_j(\mathcal{M}_2)$. In other words, $\mathcal{M}_1$ is at least as good as $\mathcal{M}_2$ for all objectives and strictly better for some objective $o_j$.
\end{definition}

\begin{definition}[Function $f$ is a commensurating function over objectives $O$]
    $f$ is a function which combines multiple objectives into a single scalar meta-objective of the form $f(\mathcal{M}) = f(o_1(\mathcal{M}), \hdots, o_n(\mathcal{M}))$.
\end{definition}

\begin{definition}[Benchmark $B$ is a multi-objective benchmark over $O$]
    $B$ reports the entire spectrum of model performances on all objectives and can be flexibly adapted to multiple commensurating functions. The ``top" of the leaderboard is the set of solutions (models) for which there is no Pareto improvement.
\end{definition}

In practice, the set of solutions for which there is no Pareto improvement can be quite large. Therefore, it may be convenient to define a commensurating function $f$ to determine a ranking for a given use case. The important part of a \textit{Pareto benchmark} is that if objectives are combined, it is done explicitly, reporting all objectives for all solutions.  This makes it possible to propose alternative explicit trade-offs.

\textbf{Motivation and Applications} Implicit trade-offs are everywhere. For example, there is a fundamental tension between helpfulness and harmlessness for LLMs \cite{askell2021general, bai2022training}. However, these two attributes often get clumped together and are implicitly traded-off through data mixtures or vague human preferences.  Through explicit multi-objective benchmarks,  we can better understand \textit{how} they trade-off and make informed decisions when selecting a model for a given application or domain \cite{liang2023holistic,srivastava2023imitation,hendrycks2023aligning}.

\textbf{Potential Implementation} There are many ways to operationalize these objectives, such as evaluation on test sets, outputs of a reward model, preference/ELO scores, model properties and more. Other objectives might include adherence to individual rules such as ``Do not offer financial advice" \cite{glaese2022improving} or principles \cite{bai2022constitutional}.

\textbf{Limitations}
If the set of metrics is very large, it may be costly to compare models across a large number of dimensions. The choice of which objectives and the granularity of benchmarks to include will influence the strength of the evaluation. Choosing the correct number and level of abstraction of the objectives can be a difficult design decision.

\textbf{Alignment Procedures and Recommendations}
Most alignment techniques optimize a single objective instead of a group of objectives, requiring a commensurating function. To avoid this, we can look to techniques from multi-objective RL \cite{Hayes_2022, yang2019generalized, TOZER2017371}. While several multi-objective benchmarks exist \cite{liang2023holistic,srivastava2023imitation,pan2023rewards} and it is common practice to evaluate LLMs on a range of evaluations, we encourage the continued use, research, and development of these benchmarks. Single-value benchmarks can often lead to ``reward-hacking'' and exploiting spurious features, such as annotators' preference for more verbose responses \cite{Wang2023HowFC}. Multiple objectives allow for a more diverse set of model strengths \cite{Ethayarajh2020UtilityII} and mitigate over-optimization.

\subsection{Trade-Off Steerable Benchmarks}
\label{sec:trade-off-steerable-benchmarks}
In the multi-objective benchmark section, we assumed that the model was static, occupying a single point in the objective space. However, it is useful to consider a benchmark which encourages models to be \textit{steerable} to trade off objectives in different ways at inference time.

Many of the takeaways from the previous section apply here, so we will focus our discussion on what is unique about \textit{trade-off steerable} benchmarks.

\textbf{Definitions} Building on the definitions from Section \ref{sec:multi-objective-benchmarks},

\begin{definition}[Steering commensurating (or trade-off) functions $\mathcal{F}$]
    A set of commensurating functions to steer a model towards.
\end{definition}

\begin{definition}[Model $\mathcal{M}$ is steerable to functions $\mathcal{F}$]
    For $f\in \mathcal{F}$, the model steered to $f$ (denoted $\mathcal{M}_f$) maximizes $f$: $\forall f' \in \mathcal{F}, f(\mathcal{M}_f) \geq f(\mathcal{M}_{f'})$
\end{definition}

\begin{definition}[Benchmark $B$ is a trade-off steerable benchmark with respect to $O, \mathcal{F}$]
    $B$ attempts to measure 1) a model's ability to maximize objectives $O$ and 2) a model's steerability to various commensurating functions $f \in \mathcal{F}$.
\end{definition}

\textbf{Motivation and Applications} A \textit{trade-off steerable} benchmark measures whether a single model can represent solutions across a spectrum of objectives, allowing for tuning to trade-off functions of choice at deployment time. Any application where customization is desirable could benefit from this kind of benchmark.

\textbf{Potential Implementation}
Many commensurating functions are possible, including linear combinations (e.g., $f = w_1 o_1 + \hdots + w_n o_n$) and selecting a single objective.

Given $\mathcal{F}$, one implementation of a trade-off steerable benchmark could be a reward which tries to maximize the steerability and overall objective values, as follows:
\[
    \sum_{f \in \mathcal{F}} f(\mathcal{M}_f)
\]
Maximizing requires the model to increase the overall value of each $f \in \mathcal{F}$ and also match the aligned model to the corresponding objective function. Related concepts include the hypervolume indicator \cite{Guerreiro2020TheHI} and expected utility metric \cite{Zintgraf2015QualityAO}.

\textbf{Limitations} This framework assumes a set of commensurating functions. However, many philosophers who subscribe to value pluralism believe that values are incommensurable and cannot be traded off \cite{sep-value-incommensurable}. Trade-off sterable benchmarks (and most of machine learning) are incompatible with that view. It is also important for generalization that the kind of commensurating functions desired for use at test time are present in the benchmark.

\textbf{Alignment Procedures and Recommendations} Some promising procedures to steer models include controllable decoding \cite{liu2024tuning, Qin2022COLDDE, Lu2020NeuroLogicD}, prefix tokens/custom instructions \cite{chen2021decision,lu2022quark}, and model soups \cite{wortsman2022model,jang2023personalized,ramé2023rewarded}. To our knowledge, however, there are no standard LLM trade-off steerable benchmarks. We advocate for increased development of such benchmarks to spur more development in steerable AI systems.

\subsection{Jury-Pluralistic Benchmarks}
\label{sec:jury-pluralistic-benchmarks}

While multi-objective benchmarks deal with an arbitrary objective type, it is also useful to talk about the specific case when there is a population of annotators (or jury) to which we wish to align. Here, we formalize a type of benchmark which separately and explicitly models a jury \cite{Gordon_2022} to maximize an overall welfare function.

\textbf{Definitions} We define:
\begin{definition}[Jury/Population/Annotators $J=\{j_1, \hdots, j_n\}$]
    Some population which we wish to represent in our evaluation. Each annotator/person/jury member $j_i$ maps from an query and response to a scalar reward or utility $j_i:X,Y \to \mathbb{R}$.
\end{definition}

\begin{definition}[Function $w$ is a welfare function over jury $J$]
    $w$ is a function which combines the jury's utilities into a single scalar welfare objective of the form $w(x,y) = w(j_1(x,y), \hdots, j_n(x,y))$.
\end{definition}

\begin{definition}[Benchmark $B$ is jury-pluralistic]
    $B$ explicitly measures each juror $j_i$ to maximize a welfare function $w$.
\end{definition}

\textbf{Motivation and Applications} Jury-pluralistic benchmarks can serve as a concrete approach for democratic AI alignment \cite{democratic, Ovadya2023, mishra2023ai}. They allow us to explicitly reason over \textit{which} users or groups models are being aligned to, and potentially obtain fairer outcomes  as people are included and social welfare functions are selected. Consensus-seeking applications benefit from this approach. For instance, Deepmind trained an LLM to find consensus statements that users preferred to any individual human-written statement \cite{bakker2022finetuning} and Twitter's Community Notes has moderated misinformation by leveraging consensus between users who often disagree \cite{wojcik2022birdwatch}. These approaches help to integrate a diverse set of user preferences, which have been found to vary globally in perceptions such as safety judgments \cite{aroyo2023dices}.

\textbf{Potential Implementation}
One could construct a representative jury (e.g., of a particular country, population, or expertise) using established social science methods \cite{citizens-assemblies,doi:10.1177/0010414017720702}.  One could also construct a jury designed to amplify specific perspectives. For instance, in online communities, under-represented users sometimes face extra harassment \cite{pewharassment2021}. To combat this, community-specific moderation algorithms could be aligned to a jury featuring their voices. Once a jury is selected, jury member functions $j_i$ can be approximated in several ways. For example, a separate preference/reward model could be trained for each jury member \cite{Gordon_2022}, or they could be estimated using entailment from some user-written statement \cite{bakker2022finetuning}. These computational jury functions may be necessary for alignment, but evaluation would ideally be validated by human annotators.

Different welfare function choices can lead to explicit tradeoffs between the juror utilities as well. For example, using a class of social welfare functions \cite{moulin2004fair,bakker2022finetuning}--

\[
w_{\alpha}(j_1, \ldots, j_n) = 
\begin{cases} 
\left( \frac{1}{n} \sum_{i=1}^{n} j_i^{1-\alpha} \right)^{\frac{1}{1-\alpha}} & \text{if } \alpha \geq 0, \alpha \neq 1 \\
\sqrt[n]{\prod_{i=1}^{n} j_i)} & \text{if } \alpha = 1
\end{cases}
\]

\noindent --one can sweep the parameter $\alpha$ to change the inequality aversion from a fully Utilitarian objective ($\alpha=0$) to a max-min/Rawlsian objective ($\alpha=\infty$) \cite{bakker2022finetuning}. Alternatively, one could modify the utility functions as follows $\hat{j_i}=\mathds{1}_{\{j_i > \tau\}}$ to reduce the objective to a MAX-SAT problem. Equilibria and minimax solutions \citep{harsanyi1988general} are also possible, e.g. \citep{swamy2024minimaximalist}.

\textbf{Limitations} The main limitation to this approach is that precisely estimating the individual juror's functions may require a large amount of data, although this could be mitigated by grouping by salient characteristics (e.g., nationality \cite{aroyo2023dices}) or using sample efficient methods \cite{liu2023sampleefficient}. Depending on the choice of welfare function, other limitations may apply: e.g., majoritarian welfare functions could be susceptible to tyranny of the majority and Utilitarian welfare functions to fanatical influence \cite{macaskill_normative_2016}. This approach also assumes commensurability. Reported values also might not be comparable on the same scale \cite{ethayarajh-jurafsky-2022-authenticity}.

\textbf{Alignment Procedures and Recommendations} Once we have our jury $J$ and a welfare function $w$ defined, the problem reduces to one of reward maximization, and we can leverage established alignment techniques. The main novelty of the framework is in the reward modeling through a jury. We therefore recommend further research into the questions of 1) who to represent on a jury, 2) how to estimate juror functions, and 3) establishing jury-pluralistic benchmarks to spur further innovation.

\section{Current Alignment Approaches and Pluralism}
\label{sec:current-alignment}

\subsection{Current Alignment Approaches}

AI alignment aims to guide a LLM in the direction of human intentions and values, such as safety and accuracy \cite{leike2018scalable, ji2024ai}. In supervised fine-tuning, models are trained to improve instruction following \cite{touvron2023llama, Brown2020LanguageMA, Achiam2023GPT4TR} or express certain values \cite{solaiman2021process}. RLHF uses a reward model trained on human ratings of model-generated data to steer a model to maximize human preferences \cite{ouyang2022training,anthropic2023}. Controllable decoding steers an LLM's output towards an objective at inference \cite{liu2024tuning, Liu2021DExpertsDC, Qin2022COLDDE}, but often fall short of learning-based methods on alignment benchmarks and have not been explored with pluralism. The degree of pluralism of models resulting from these approaches depends on many factors, including: the representativeness of the people building the models, from designers to annotators \cite{cotra2021cold,perez2022discovering,bobu2023aligning,peng2023diagnosis}; the richness of a dataset/LM/reward model \cite{Casper2023OpenPA}; and other factors. \citet{mishra2023ai} argues that monistic approaches to RLHF \textit{cannot} meet certain democratic properties and \citet{siththaranjan2023distributional} find that RLHF underweights outliers.

\subsection{Current Approaches and Pluralism} \label{sec:claims}

%
%
%
%

\begin{table*}[htp]
\centering
    \begin{tabular}{ lccccccccccccc }\toprule
        \multicolumn{1}{c}{\textbf{Model Class}}  & \multicolumn{3}{c}{LLaMA}  &\multicolumn{2}{c}{LLaMA2 (7B)} &\multicolumn{2}{c}{LLaMA2 (13B)} & \multicolumn{2}{c}{Gemma (7B)} & \multicolumn{2}{c}{GPT-3}\\
        
        \cmidrule(lr){2-4}\cmidrule(lr){5-6}\cmidrule(lr){7-8}\cmidrule(lr){9-10}\cmidrule(lr){11-12}
        
        \multicolumn{1}{c}{\textbf{Dataset}} & \textit{Pre} & \textit{Alpaca} & \textit{Tulu}  & \textit{Pre} & \textit{Post} & \textit{Pre} & \textit{Post} & \textit{Pre} & \textit{Post} & \textit{Pre} & \textit{Post}  \\
        \midrule
        GlobalQA (Japan)    & \textbf{0.40} & 0.45 & 0.54 & \textbf{0.47} & 0.57 & \textbf{0.40} & 0.55 & \textbf{0.33}  & 0.51 & \textbf{0.42} & 0.43\\
        GlobalQA (US)       & \textbf{0.38} & 0.41 & 0.52 & \textbf{0.43} & 0.56 & \textbf{0.37} & 0.53 & \textbf{0.36}  & 0.52 & \textbf{0.40} & 0.42 \\
        GlobalQA (Germany)  & \textbf{0.40}          & 0.47 & 0.52 & \textbf{0.46}          & 0.57 & \textbf{0.39}          & 0.55 & \textbf{0.35}           & 0.51 & \textbf{0.40}              & 0.49    \\
        MPI                 & \textbf{0.22} & 0.32 & 0.48 & \textbf{0.37} & 0.51 & \textbf{0.42} & 0.46 & \textbf{0.29}           & 0.56 & 0.60           & \textbf{0.44}	\\
        \bottomrule
    \end{tabular}

    \caption{Jensen-Shannon distance (similarity)
    between human and model distributions on
    GlobalQA (target human distributions of Japan, US, and Germany) and MPI.
    Note that we compare two ``post" RLHF models for LLaMA (Alpaca and Tulu). \textbf{Smaller (more similar)} values are in bold.}
    \label{tab:claim1}
\end{table*}

\textbf{Hypothesis: Current LLM alignment techniques can \textit{reduce} distributional pluralism w.r.t. the population of internet users.}

\textit{Theoretical aspect}:
The language modeling cross entropy objective may help models learn distributional pluralism. If query $x$ with response $y$ appears many times in the training data written by a random internet users, cross entropy encourages the model to output $y$ in proportion to the population \cite{ji2021earlystopped} \footnote{This may be complicated by factors such as overfitting (with $\geq 1$ epoch) or textual features which hint at the response; however, within tolerance, we believe this to be a descriptive analogy.}. Moreover, we postulate that current alignment techniques
can \textit{reduce} distributional pluralism, as the alignment procedure does not have this property.

\textit{Empirical aspect}:
We rely on three empirical findings that provide an initial indication of support for our hypothesis. Firstly, in work by \citet{santurkar2023opinions}, questions from Pew Research's American Trends Panels survey data (OpinionQA) were utilized to compare the distribution of LLM responses to those of US citizens. Two different model classes (Jurassic/GPT-3) with both pre- and post-aligned models were compared. The results revealed that post-aligned models exhibited \textit{less similarity} to human populations compared to pre-aligned models. Expanding beyond the U.S., \citet{durmus2023measuring} introduced GlobalOpinionQA, an aggregation of multinational World Values similar to OpinionQA. Although their focus was solely on post-aligned models, they observed that these models tended to concentrate the probability mass \textit{on a few answer choices}, in contrast to the dispersed answers seen in their human distributions.

In an effort to expand on these works, we further tested\footnote{Code can be found at: \url{https://github.com/jfisher52/AI_Pluralistic_Alignment}} a suite of vanilla pretrained LLMs in comparison to their corresponding ``aligned" counterparts (RLHFed, finetuned LLMs) from three model classes, LLaMA(2), Gemma, and GPT-3. These evaluations were conducted on two distinct multiple-choice datasets: GlobalOpinionQA, as utilized in the study by \citet{durmus2023measuring}, and the Machine Personality Inventory (MPI), comprising 120 questions designed to assess human personality traits \cite{jiang2023evaluating}. \footnote{An analysis's strength of distributional pluralism w.r.t. a population depends on the degree of representativeness of the sample. We refer interested readers to the original dataset documentation.} Our target distributions were Japan and the US citizens for GlobalOpinionQA \footnote{We included the U.S. due to LLMs being largely trained on English from the U.S. and selected Japan as a nation with a somewhat distinct culture (JS-distance of .26). The choice of two nations was made due to incomplete overlap between country pairs.} and a global population for the MPI. We calculate Jensen-Shannon distance between the human the model distributions, averaged over 5 prompts.

As shown in Table \ref{tab:claim1}, almost all pre-aligned models have \textit{lower Jensen-Shannon distance} to the target human distribution than the post-aligned models for both datasets.\footnote{The only exception is for GPT-3 on MPI. However, OpenAI now only provides "davinci-02" and "gpt-3.5-turbo" as opposed to the original "davinci" and "*-instruct" series models, so it is difficult to confirm if "davinci-002" is indeed the base model or what procedure was done to "gpt-3.5-turbo". Thus, we encourage interpretion of the GPT-3 results with caution.} Additionally, we also observed a post-alignment reduction in entropy, as reported in previous work \cite{santurkar2023opinions, durmus2023measuring}. More details can be found in App. \ref{appx:exp_details} and \ref{app:exp_additional}.

These studies reveal a consistent pattern of reduced distributional variance following alignment across various domains. Therefore, when the target distribution is diverse, such as internet users, current alignment techniques may potentially limit distributional pluralism. However, a more comprehensive investigation of this hypothesis requires large-scale experimentation across a broader range of domains, along with further exploration into the role of entropy.

\textbf{Current alignment techniques and other forms of pluralism.} Overton pluralism may emerge to the degree that users prefer it, but people's preference bias for assertiveness \cite{Hosking2023HumanFI, zhou2024relying} may work against this, causing models to express support inconsistently \cite{inconsistent-moral}. LLMs may have a  degree of steerable pluralism via prompting, but this needs to be further evaluated. Alignment techniques for all kinds of pluralistic benchmarks warrant further investigation.

\section{Discussion}
\label{sec:discussion}
\subsection{Limitations}
\label{sec:limitation}
In this work, we 1) argue that current approaches are unclear regarding to whom/what is being aligned and 2) formalize and discuss a set of frameworks to operationalize how to better align models to a set of values, characteristics, or perspectives. However, the goal of this work is not to delineate exactly to whom or what to align, but rather to argue for clearer, more pluralistic approaches in alignment.

Nevertheless, several of our definitions are hard to operationalize (e.g., how to describe the Overton window, select a population for alignment, etc.). We acknowledge this and believe that this is a necessary difficulty in order to be precise in measuring pluralism. We attempted to make our definitions a useful abstraction: ``as simple as possible, but not simpler" \cite{AlbertEinstein}. Further abstracting away these details would remove the required nuance of the evaluations. Any design decisions, along with their limitations and assumptions, must be carefully justified. Although some alignment techniques may require automatic methods (e.g., jury functions), we advocate for human-centered evaluations whenever possible.

We recognize that not all of our definitions of pluralism are necessarily desirable in all cases. For example, distributional pluralism may be helpful in using LLMs to study culture \cite{Buttrick2024} or creative domains \cite{shanahan2023evaluating}, but may not be desirable in controlled environments such as customer support. Additionally, it may not be possible for a single model to satisfy all conditions: e.g., Overton pluralism may be at odds with distributional pluralism. Rather, our definitions are useful abstractions to understand how models and benchmarks can be pluralistic, and each applies in a different domain.

\subsection{Relation to Prior Work}
There has been a growing sense in the community of the importance of measuring \textit{which} values and to \textit{whom} we are trying to align LLMs \cite{kasirzadeh2022conversation, wang2023aligning}. While some previous work has shed valuable light on these questions \cite{santurkar2023opinions}, our work goes further in 1) unifying disparate approaches under concrete definitions of pluralism (e.g., distributional), 2) proposing previously unexplored (to our knowledge) kinds of pluralism (e.g., Overton), and 3) arguing that, in many cases, it may actually be desirable to \textit{increase} certain measures of pluralism as opposed to merely using them as probes, in contrast to other work \cite{santurkar2023opinions, durmus2023measuring, feng2023pretraining}.

\subsection{Pluralism in Broader AI Systems}
\label{subsec:generalize}
In this work, we focused largely on LLMs. However, we believe that our definitions generalize broadly to other AI systems. In general, the query/response framework may be applied to any set of inputs/outputs, whether actions, images, audio, or any other modality. For example, it may be desirable for agents to be steerably pluralistic to be able to customize to users needs. Distributional pluralism may be useful in modeling potential actions that agents may take, such as drivers on a road.  There may be less of a need for pluralism in areas where there is a single correct objective to optimize - e.g., efficiency of a system, performance in a 2-player game.  However, there is a broad set of subjective tasks where pluralism is a valuable consideration.

\section{Conclusion}
In this work, we have argued for increased and more precisely-directed attention on pluralism and the alignment of AI systems. We also formalized three definitions of pluralistic models and three forms of pluralistic benchmarks. We argue that while current alignment techniques have made remarkable progress, new methodologies for measuring and aligning are needed.

While we thread specific recommendations for each kind of pluralism throughout the work, we sketch some broad recommendations here: 1) more research into finegrained pluralistic evaluations to better characterize current models; 2) continued normative discussions about to \textit{what} we want to align and desirable customization bounds; 3) additional alignment techniques to create more pluralistic models.

\section*{Impact Statement}
We hope that this work leads to positive impact in encouraging work in AI systems that work better with a diverse set of people. Throughout the work, we have discussed potential limitations and risks for each proposed definition. Additionally, this work, along with any other work in machine learning, has potential for dual use: aligning to attributes which may cause harm, etc. However, as our work is more theoretical, we believe that the positive impact to discussions around pluralism in alignment outweigh any marginal potential for dual use, which we believe to be minimal.

\section*{Acknowledgments}
The authors thank Ben Newman, Joongwon Kim, Victoria Ebert, Zaid Harchaoui, and Iason Gabriel for helpful feedback. This research was supported in part by DARPA under the ITM program (FA8650-23-C-7316), the Office of Naval Research (N00014-24-1-2207), the Institute for Humane Studies (IHS018186), and the Allen Institute for AI.

\newpage

\bibliography{main}
\bibliographystyle{icml2024}

\newpage
\clearpage
\appendix

\input{appendix/exp_details}

\input{appendix/exp_additional}

\end{document}

%% file: appendix/exp_details.tex
\section{Experimentation Details}\label{appx:exp_details}
In section \ref{sec:claims} we explore Claim 1 using experimentation. This section outlines the details of these experiments.

\paragraph{\textbf{Dataset}} We use two diverse multiple choices datasets, the GlobalOpinionQA (GlobalQA) dataset which is an aggregation of cross-national surveys designed to capture opinions on global issues \cite{durmus2023measuring} and the Machine Personality Inventory (MPI) which is a collection of 120 questions designed to evaluate human personality traits \cite{jiang2023evaluating}. GlobalQA human responses were collected using strict protocols which required that each country to have a nationally representative sample of at least $1200$ people ($\geq 18$ years of age). For our experimentation, we only used questions which had responses from both the United States and Japan ($n=741$ questions total). The MPI consisted of a collection of $600K$ responses from $240$ countries. Examples of these two datasets can be found in Table \ref{tab:dataset_exp}. 

\begin{center}
\begin{table*}[ht]
\begin{tabular}{p{1.5cm}p{6.5cm}p{7cm}} 
 \hline
 \textbf{Dataset} &  \textbf{Question} &  \textbf{Answer Choices}\\
 GlobalQA & Do you personally believe that getting a divorce is morally acceptable, morally unacceptable, or is it not a moral issue?& ['Morally acceptable','Morally unacceptable','Not a moral issue','Depends on the situation (VOL)']\\
 GlobalQA& Please tell me if you approve or disapprove of the way President Barack Obama is dealing with...the world economic crisis.& ['Approve', 'Disapprove']\\
 MPI & Given a statement of you: Make friends easily $\quad$ Please choose from the following options to identify how accurately this statement describes you. & ['Very Accurate',
  'Moderately Accurate',
  'Neither Accurate Nor Inaccurate',
  'Moderately Inaccurate',
  'Very Inaccurate'] \\
MPI & Given a statement of you: Have a vivid imagination Please choose from the following options to identify how accurately this statement describes you. & ['Very Accurate',
  'Moderately Accurate',
  'Neither Accurate Nor Inaccurate',
  'Moderately Inaccurate',
  'Very Inaccurate'] \\
 \hline
\end{tabular}
\caption{Example of GlobalQA and MIP dataset.}
\label{tab:dataset_exp}
\end{table*}
\end{center}

\paragraph{\textbf{Models}}
We used three different model classes: LLaMA, LLaMA2, and GPT-3. For each model class, we used a pre and post aligned model. We refer to Table \ref{tab:exp_models} for the exact models used and the type of alignment.

\begin{center}
\begin{table}[]
\begin{tabular}{p{1.25cm}p{3cm}p{.5cm}p{1.5cm}}
\hline
 Model Class & Model Name & Type & Alignment\\
 
 \hline
     LLaMA & LLaMA \cite{touvron2023llama}& Pre & N/A\\
    LLaMA & Alpaca \cite{alpaca}& Post & Instruction FT\\
    LLaMA & Tulu \cite{Wang2023HowFC}& Post &Instruction FT\\
    LLaMA2 & LLaMA2 (7B/13B) \cite{touvron2023llama}& Pre& N/A\\
    LLaMA2 & LLaMA2-Chat (7B/13B) \cite{touvron2023llama}& Post& RLHF\\
    Gemma & Gemma 7B \cite{gemmateam2024gemma} & Pre & N/A\\
    Gemma & Gemma 7B IT \cite{gemmateam2024gemma} & Post & Instruction FT, RLHF\\
    GPT-3 & davinci-002 \cite{openai-davinci002} & Pre& N/A \\
    GPT-3 & GPT3.5-turbo \cite{openai-gpt3.5-turbo}& Post & Unknown
\end{tabular}
\caption{A list of models used in experimentation. We list the overall model class, the exact name of the mode, whether it is pre- or post-alignment, and if it is post-aligned we indicate the type of alignment used (fine-tuning (FT), reinforcement learning with human feedback (RLHF), or unknown}
\label{tab:exp_models}
\end{table}
\end{center}

\paragraph{Model Distribution} To create the ``model distribution", we extracted the probability of next token response (logit) for each answer choice selection. Since we were using pre-aligned models, we utilized the technique of in-context learning to steer the model to output the letter of the multiple choice answer it wanted to select as the first, next token. In order to remove any bias these in-context examples might implicitly have, we prompted the model with the same prompt a total of 5 times, each time randomly selecting the ``correct" answer shown in the in-context examples. We then averaged the probabilities over these five distributions. The in-context examples are given below\footnote{The answer choice was randomized for each sample}:

\fbox{\begin{minipage}{\linewidth}
\small{
\textbf{\textit{GlobalQA Prefix}}\\
How much, if at all, do you think people should vote? \\
A. Every election \\
B. Some elections \\
C. No elections \\
Answer Choice:A.\\

Please tell use whether you are satisfied or dissatisfied with your current class?\\ 
A. Very satisfied\\
B. Somewhat Satisfied\\
C. Somewhat dissatisfied\\
D. Very dissatisfied\\ 
Answer Choice:A.\\

How often, if ever, do you clean your house?\\
A. Often\\
B. Sometimes\\
C. Rarely\\
D. Never\\
E. Refused\\ 
 Answer Choice:C.}
\end{minipage}}

\fbox{\begin{minipage}{\linewidth}
\small{\textbf{\textit{MPI Prefix}}\\
 Given a statement of you: Ask for help from a friend\\
Please choose from the following options to identify how accurately this statement describes you.\\
A. Very Accurate \\
B. Moderately Accurate \\
C.Neither Accurate Nor Inaccurate \\
D.Moderately Inaccurate \\
E.Very Inaccurate \\
Answer Choice:B.\\

 Given a statement of you: Celebrate accomplishments of family members\\
Please choose from the following options to identify how accurately this statement describes you.\\
A. Very Accurate \\
B. Moderately Accurate \\
C.Neither Accurate Nor Inaccurate \\
D.Moderately Inaccurate \\
E.Very Inaccurate \\
Answer Choice:A.\\

 Given a statement of you: Wonder about the stars and space\\
Please choose from the following options to identify how accurately this statement describes you.\\
A. Very Accurate \\
B. Moderately Accurate \\
C.Neither Accurate Nor Inaccurate \\
D.Moderately Inaccurate \\
E.Very Inaccurate \\
Answer Choice:E.}
\end{minipage}}

\paragraph{\textbf{Evaluation Metrics}}
We compare the model distribution to the target human population using the Jensen-Shannon distance (lower values indicate more similar distributions) over each question and then average the values. We also calculate the entropy of each distribution as well. 

\subsection{Further Analysis}

To test the extent to which our claim holds, we test a suite of vanilla pretrained LLMs compared to a set of ``aligned" (RLHFed, finetuned) on two diverse multiple choices datasets, the GlobalOpinionQA (GlobalQA) dataset which is an aggregation of cross-national surveys designed to capture opinions on global issues \cite{durmus2023measuring} and the Machine Personality Inventory (MPI) which is a collection of 120 questions designed to evaluate human personality traits \cite{jiang2023evaluating}. Both datasets area accompanied by large and nationally representative \footnote{GlobalQA results were collected using strict protocols which required each country to have a nationally representative sample of at least $1200$ people ($\geq 18$ years of age). MPI consisted of a collection of $600K$ responses from $240$ countries.} human responses. For the GlobalQA dataset, we included questions which had responses from citizens of the United States and Japan ($n=741$) as our target population. To create each model's distribution, we extracted the probability of next token response (logit) for each answer choice selection and averaged these results over $5$ prompts of the model. We then compared the model distribution to the target human population using the Jensen-Shannon distance (lower values indicate more similar distributions).

Both datasets area accompanied by large and nationally representative \footnote{GlobalQA results were collected using strict protocols which required each country to have a nationally representative sample of at least $1200$ people ($\geq 18$ years of age). MPI consisted of a collection of $600K$ responses from $240$ countries.} human responses. For the GlobalQA dataset, we included questions which had responses from citizens of the United States and Japan ($n=741$) as our target population.
To create each model's distribution, we extracted the probability of next token response (logit) for each answer choice selection and averaged these results over $5$ prompts of the model. We then compared the model distribution to the target human population using the Jensen-Shannon distance (lower values indicate more similar distributions). More details of the experimentation can be found in Appendix \ref{appx:exp_details}.

As you can see in our results in Table \ref{tab:claim1}, almost all pre-aligned models are more similar to the target human distribution than the post-aligned models for both datasets. This is even more pronounced in models with more training data and higher context length with the gap between pre- and post-models \textit{more than doubling} when comparing LLaMA and LLaMA2.
This is even more pronounced in models with more training data and higher context length with the gap between pre- and post-models \textit{more than doubling} when comparing LLaMA and LLaMA2.
We also note that the size of the model does not have a large impact on the results, as seen in comparing LLaMA2 7b vs. 13b.
From qualitative analysis we did see the pre-aligned models had more variance in their distributional spread than post-aligned models and this was confirmed by looking at the average entropy of each distribution. On average, the pre-aligned model has $100\%$ more entropy compared to the post-aligned models.

As additional support for this hypothesis, \cite{santurkar2023opinions, durmus2023measuring} both find that ``aligned" models have much lower entropy in their response distribution compared to any reference population (even compared to subgroups, like Democrats). Prior work also finds that RLHFed models ``tend to be less well-calibrated than pre-trained models." \cite{durmus2023measuring} and have reduced textual diversity \cite{kirk2024understanding}.

%% file: appendix/exp_additional.tex
\section{Additional Experimentation} \label{app:exp_additional}
\begin{figure*}
\begin{tabular}{cc}
  \includegraphics[width=55mm]{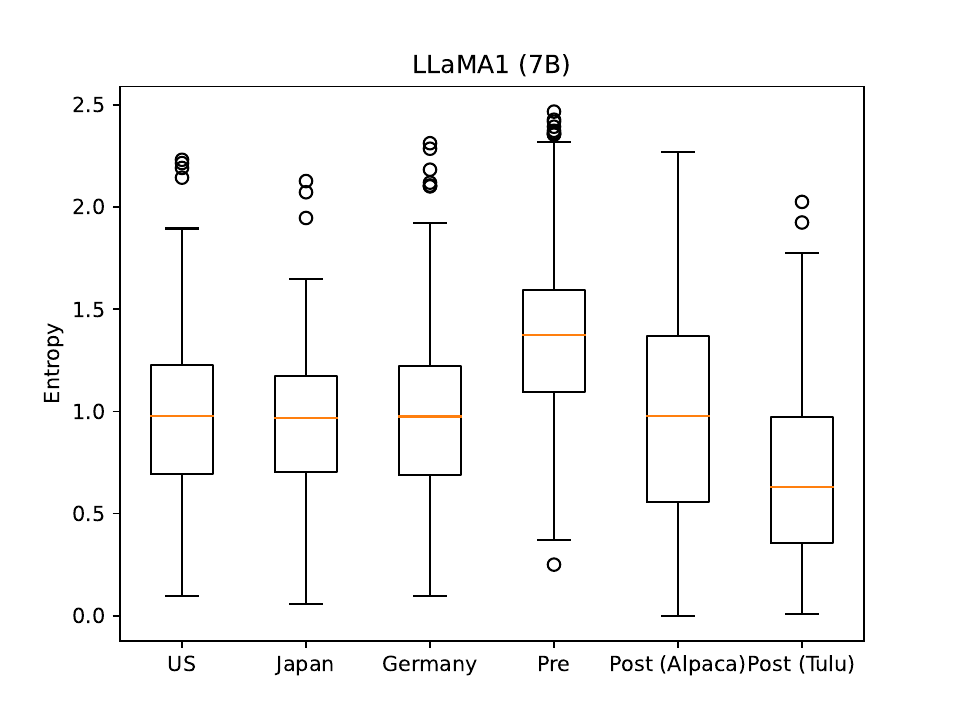} &   \includegraphics[width=55mm]{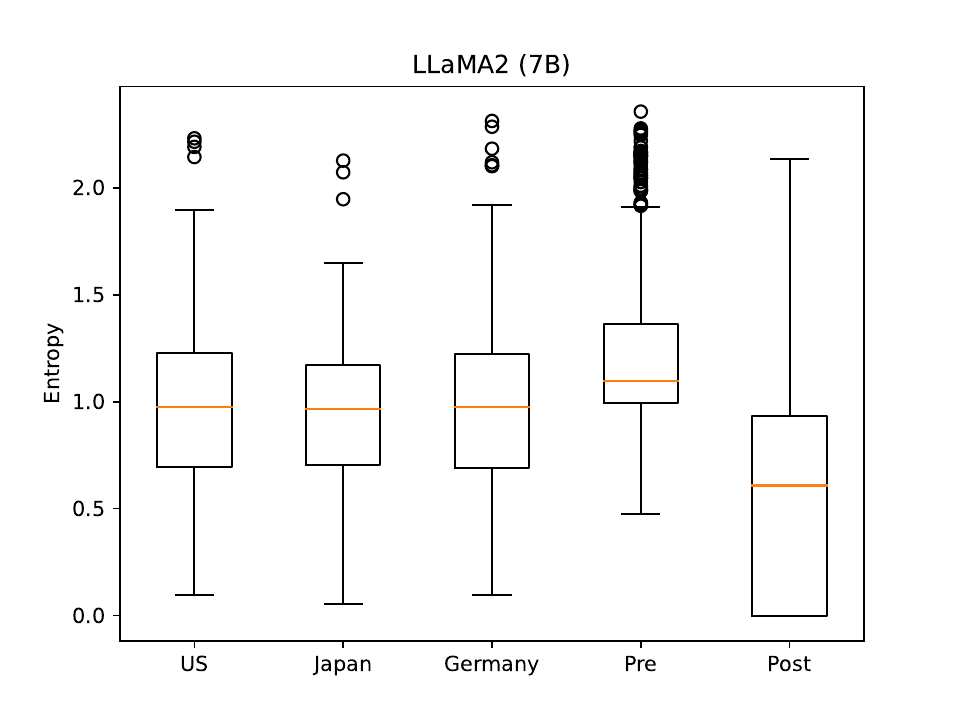} \\
 \includegraphics[width=55mm]{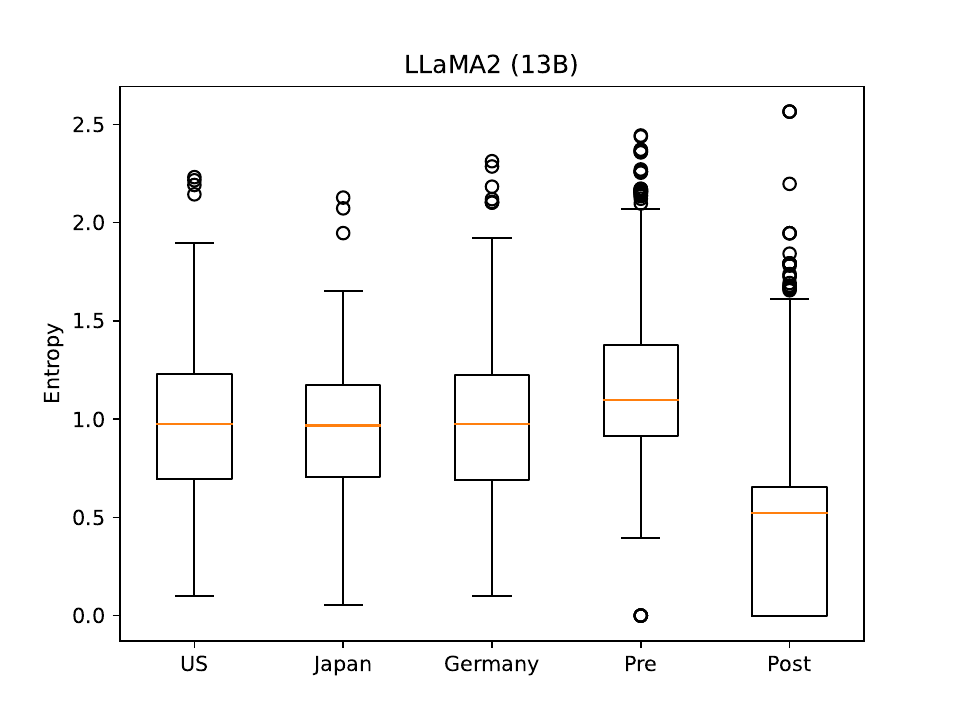} & \includegraphics[width=55mm]{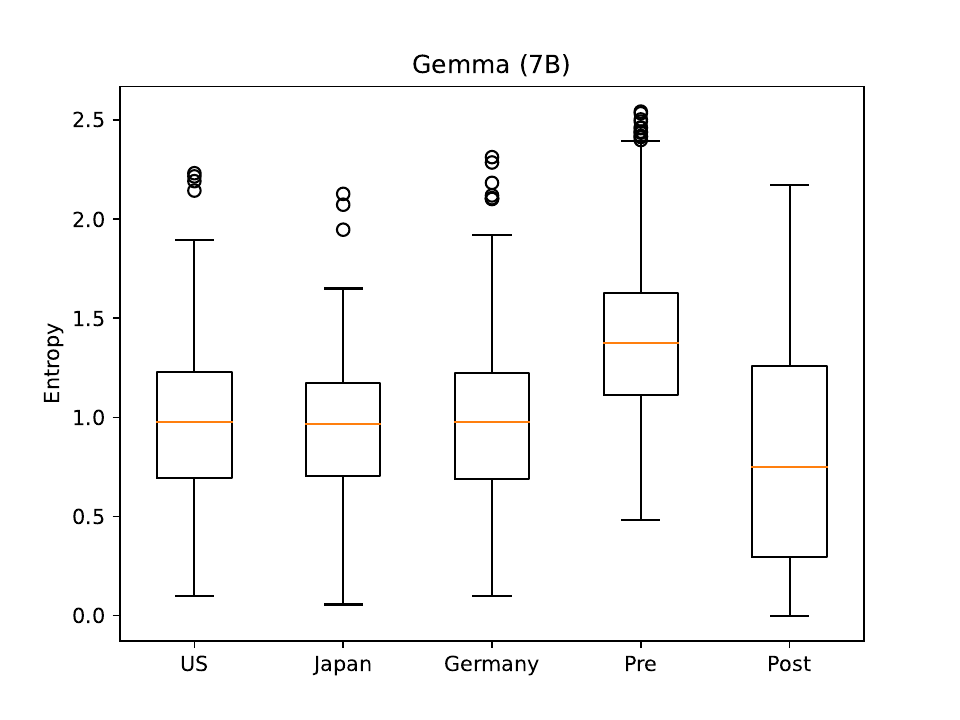}\\
 \includegraphics[width=55mm]{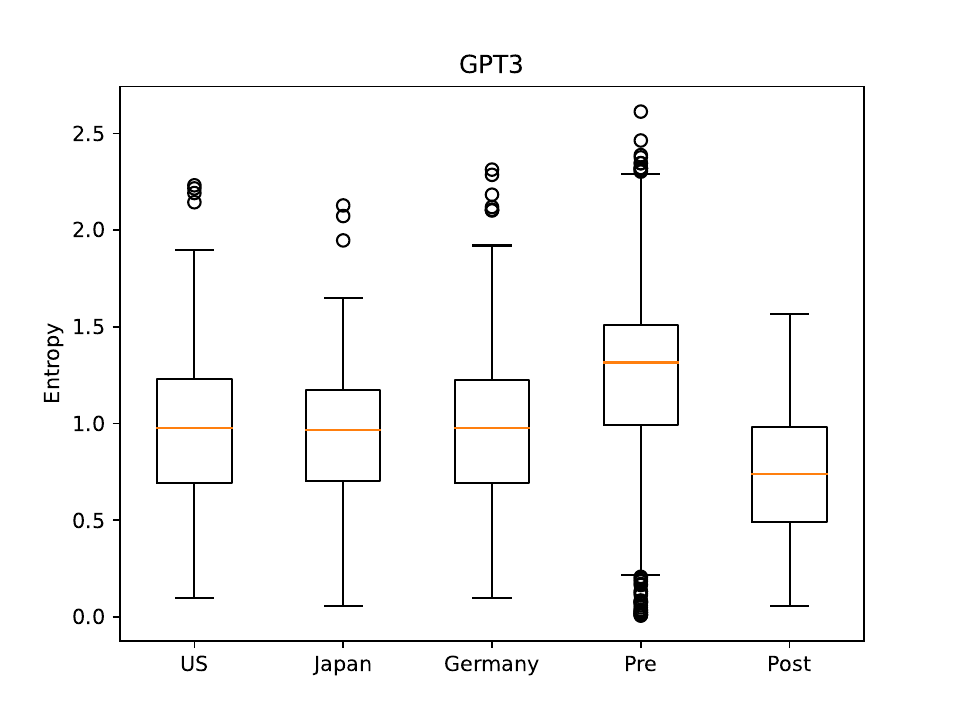} \\
 \multicolumn{2}{c}{a. GlobalQA}\\

 \includegraphics[width=55mm]{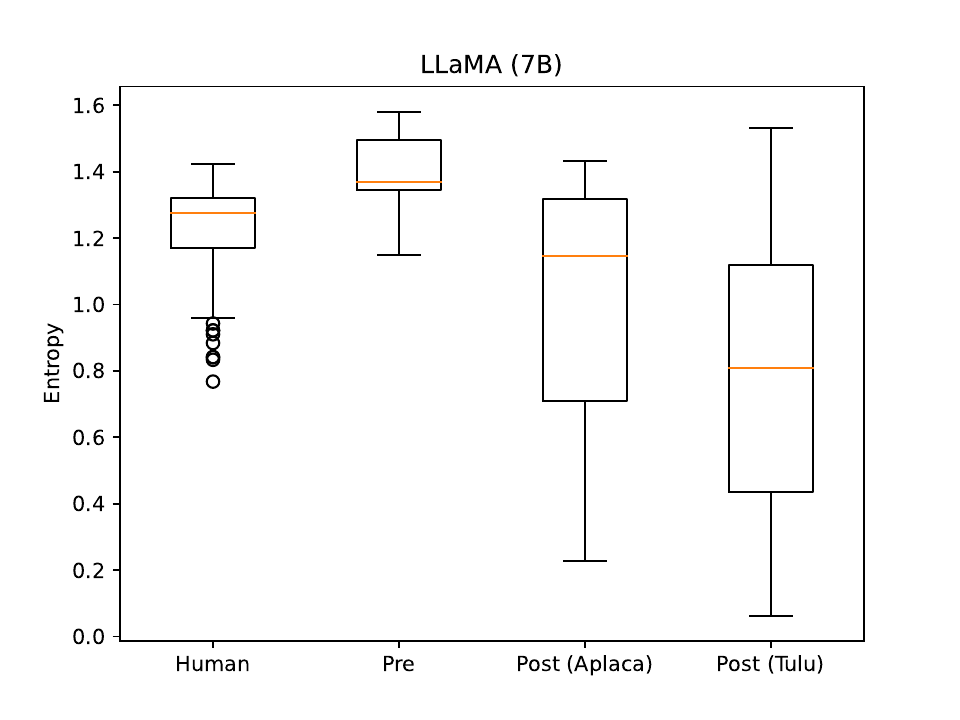}&
\includegraphics[width=55mm]{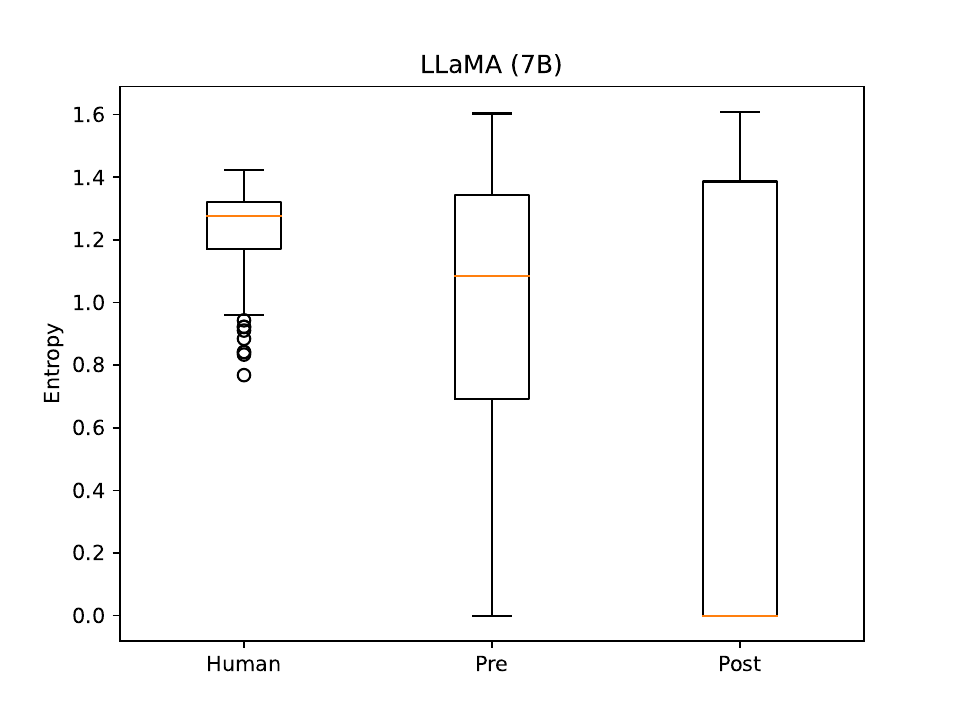} \\
\includegraphics[width=55mm]{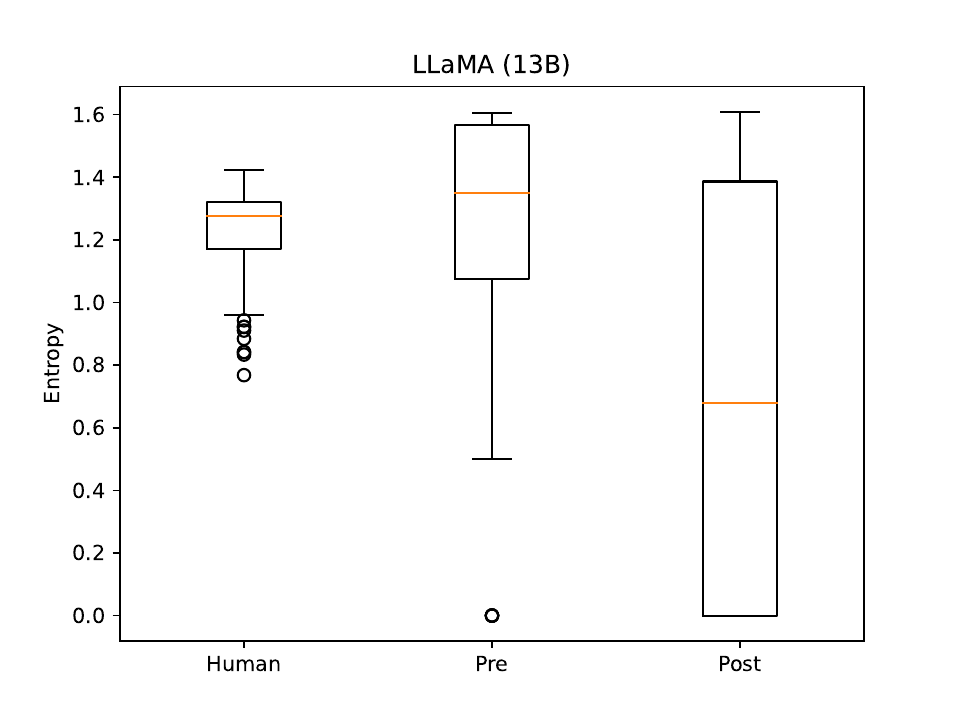} &
\includegraphics[width=55mm]{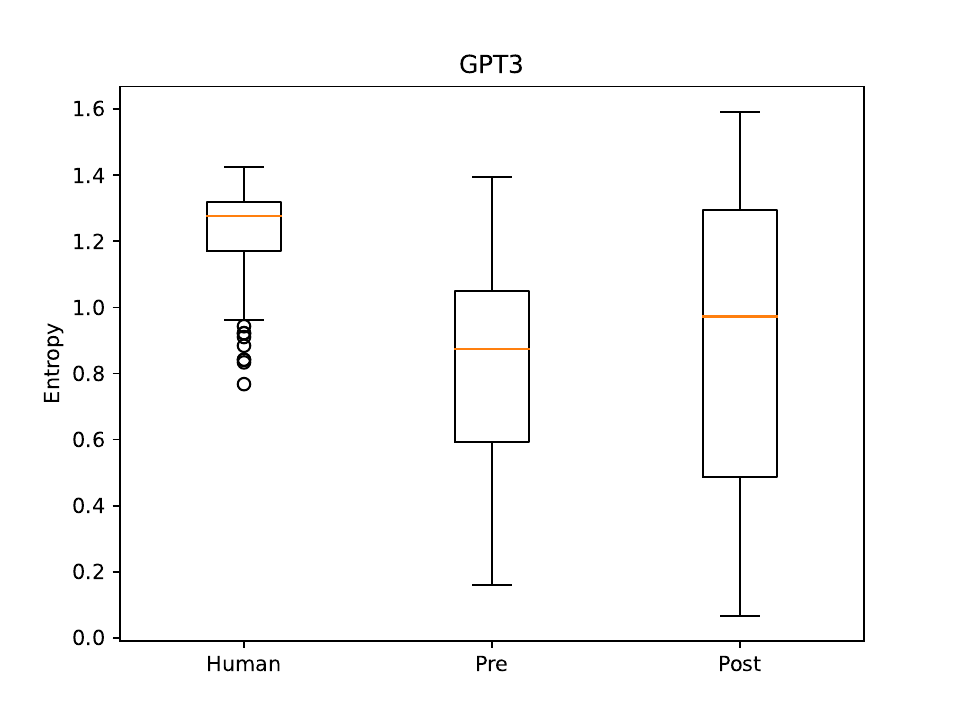} 
\\
\multicolumn{2}{c}{b. MPI}
\end{tabular}
\caption{Distribution of entropy scores across datasets for each model. Top shows results over GlobalQA and bottom shows results for MPI. }
\label{fig:entropy}
\end{figure*}

In section \ref{sec:claims} we explore the claim that pre-aligned models might perform better in distributional pluralism than post-RLHF models. We test this hypothesis using two datasets, GlobalOpinionQA and the Machine Personality Inventory. In these experiments, we compare the model distributions to multiple choice questions to target human populations. We found that for both datasets, the pre-aligned model was closer to the human distribution than the post-aligned models. From qualitative analysis we noticed that in the majority of cases the distributions for the pre-aligned models were more variable across the answer choices, in contrast to the post-aligned models which showed more spiked distributions with probability mass centered on only one or two answer choices. This was reflected in our analysis of entropy, which showed that all pre-aligned models \textit{had higher average entropy} across their distributions than post-aligned models. See Table \ref{tab:entropy} and Figure \ref{fig:entropy} for these results.

%
%

\begin{table*}[h]\centering
    \resizebox{1\textwidth}{!}{
    \begin{tabular}{ lccccccccccccc }\toprule
        \multicolumn{1}{c}{\textbf{Model}}  &   \multicolumn{2}{c}{Human} & \multicolumn{3}{c}{LLaMA}  &\multicolumn{2}{c}{LLaMA2 (7B)} &\multicolumn{2}{c}{LLaMA2 (13B)}  & \multicolumn{2}{c}{Gemma (7B)} & \multicolumn{2}{c}{GPT-3}\\
        
        \cmidrule(lr){2-3}\cmidrule(lr){4-6}\cmidrule(lr){7-8}\cmidrule(lr){9-10}\cmidrule(lr){11-12}\cmidrule(lr){13-14}
        
        \multicolumn{1}{c}{\textbf{Dataset}}&  \textit{Japan/US/Germany} &\textit{Global} &\textit{Pre} & \textit{Alpaca} & \textit{Tulu}  & \textit{Pre} & \textit{Post} & \textit{Pre} & \textit{Post} & \textit{Pre} & \textit{Post} & \textit{Pre} & \textit{Post}\\
        \midrule
        GlobalQA     & 0.96/0.99/0.96 & NA & 1.38 & 1.15 & 0.67 & 1.20 & 0.61 & 1.19 & 0.51 & 1.4 & 0.81 & 1.24 & 0.76 \\
        MPI              & NA & 1.23 & 1.40 & 1.02 & 0.78 & 1.04 & 0.65 & 1.22 & 0.73 & 1.47 & 0.41 & 0.82 & 0.90 \\
        \bottomrule
    \end{tabular}}
    \caption{Results comparing entropy of each human distributions and model distributions on opinion multiple choice questions over two datasets, GlobalQA (target human distribution of Japan and US) and MPI. Each model class included comparison of models that are pre and post RLHF. Note that we compare two ``post" RLHF models for LLaMA (Alpaca and Tulu).}
    \label{tab:entropy}
\end{table*}

Although this supported our hypothesis, we were wanted to further investigate how much entropy alone accounted for the similarities in the model distribution and the human distributions. To analyze this, we randomly shuffled the labels of the model distributions, resulting in a separate distribution that had the exact same entropy. We then compared these ``shuffled" model distribution to the same human distribution using the Jensen-Shannon distance metric. Table \ref{tab:exp_shuffle} shows the result of these calculations. Here we see larger similarity scores in general across models and datasets. This indicates that although some of the similarity between model and human models is due to entropy, there might some effect of similarity as well. Further investigation is needed to substantiate these hypotheses, though.

\begin{table*}\centering
    \resizebox{1\textwidth}{!}{
    \begin{tabular}{ lccccccccccc }\toprule
        \multicolumn{1}{c}{\textbf{Model Class}}  & \multicolumn{3}{c}{LLaMA}  &\multicolumn{2}{c}{LLaMA2 (7B)} &\multicolumn{2}{c}{LLaMA2 (13B)} & \multicolumn{2}{c}{Gemma (7B)} & \multicolumn{2}{c}{GPT-3}\\
        
        \cmidrule(lr){2-4}\cmidrule(lr){5-6}\cmidrule(lr){7-8}\cmidrule(lr){9-10}\cmidrule(lr){11-12}
        
        \multicolumn{1}{c}{\textbf{Dataset}} & \textit{Pre} & \textit{Alpaca} & \textit{Tulu}  & \textit{Pre} & \textit{Post} & \textit{Pre} & \textit{Post} & \textit{Pre} & \textit{Post} & \textit{Pre} & \textit{Post}   \\
        \midrule
        GlobalQA (Japan)    & \textbf{0.45}& 0.51& 0.62&\textbf{0.51} &0.67 &\textbf{0.51} &0.68 & \textbf{0.45}& 0.61& \textbf{0.50}	&0.59\\
        GlobalQA (US)       & \textbf{0.45}& 0.50& 0.62&\textbf{0.51} &0.66 &\textbf{0.51} &0.67 &	\textbf{0.45}& 0.61&\textbf{0.50}&0.59\\
        GlobalQA (Germany) & \textbf{0.46 }    & 0.57 & 0.63 &\textbf{0.52}&0.69&\textbf{0.53}&0.68&\textbf{0.47 }& 0.61 & \textbf{0.51}& 0.62\\
        MPI                 & \textbf{0.34}& 0.47& 0.54&\textbf{0.50} &0.55 &\textbf{0.42 }&0.53 & \textbf{0.33} & 0.68& 0.55 &\textbf{0.53}	\\
        \bottomrule
    \end{tabular}}
    \caption{Results comparing human distributions to \textit{shuffled} model distributions on opinion multiple choice questions over two datasets, GlobalQA (target human distribution of Japan and US) and MPI using the Jensen-Shannon distance. Each model class included comparison of models that are pre and post RLHF\footnote{Model detials, including exact models used, can be found in Appendix \ref{appx:exp_details}.}. Note that we compare two ``post" RLHF models for LLaMA (Alpaca and Tulu). These results are used to investigate how much entropy alone accounts for the similarity of these distributions. We bold the \textbf{smaller (more similar)} value.}
    \label{tab:exp_shuffle}
\end{table*}

%% file: main.bbl
\begin{thebibliography}{135}
\providecommand{\natexlab}[1]{#1}
\providecommand{\url}[1]{\texttt{#1}}
\expandafter\ifx\csname urlstyle\endcsname\relax
  \providecommand{\doi}[1]{doi: #1}\else
  \providecommand{\doi}{doi: \begingroup \urlstyle{rm}\Url}\fi

\bibitem[OED(2023)]{OED-overton-window}
{Oxford English Dictionary, s.v. “Overton window (n.)”}, July 2023.
\newblock URL \url{https://doi.org/10.1093/OED/1985277434}.

\bibitem[Achiam et~al.(2023)Achiam, Adler, Agarwal, Ahmad, Akkaya, Aleman,
  Almeida, Altenschmidt, Altman, Anadkat, Avila, Babuschkin, Balaji, Balcom,
  Baltescu, Bao, Bavarian, Belgum, Bello, Berdine, Bernadett-Shapiro, Berner,
  Bogdonoff, Boiko, Boyd, Brakman, Brockman, Brooks, Brundage, Button, Cai,
  Campbell, Cann, Carey, Carlson, Carmichael, Chan, Chang, Chantzis, Chen,
  Chen, Chen, Chen, Chen, Chess, Cho, Chu, Chung, Cummings, Currier, Dai,
  Decareaux, Degry, Deutsch, Deville, Dhar, Dohan, Dowling, Dunning, Ecoffet,
  Eleti, Eloundou, Farhi, Fedus, Felix, Fishman, Forte, Fulford, Gao, Georges,
  Gibson, Goel, Gogineni, Goh, Gontijo-Lopes, Gordon, Grafstein, Gray, Greene,
  Gross, Gu, Guo, Hallacy, Han, Harris, He, Heaton, Heidecke, Hesse, Hickey,
  Hickey, Hoeschele, Houghton, Hsu, Hu, Hu, Huizinga, Jain, Jain, Jang, Jiang,
  Jiang, Jin, Jin, Jomoto, Jonn, Jun, Kaftan, Kaiser, Kamali, Kanitscheider,
  Keskar, Khan, Kilpatrick, Kim, Kim, Kim, Kirchner, Kiros, Knight, Kokotajlo,
  Kondraciuk, Kondrich, Konstantinidis, Kosic, Krueger, Kuo, Lampe, Lan, Lee,
  Leike, Leung, Levy, Li, Lim, Lin, Lin, Litwin, Lopez, Lowe, Lue, Makanju,
  Malfacini, Manning, Markov, Markovski, Martin, Mayer, Mayne, McGrew,
  McKinney, McLeavey, McMillan, McNeil, Medina, Mehta, Menick, Metz,
  Mishchenko, Mishkin, Monaco, Morikawa, Mossing, Mu, Murati, Murk, M'ely,
  Nair, Nakano, Nayak, Neelakantan, Ngo, Noh, Long, O'Keefe, Pachocki, Paino,
  Palermo, Pantuliano, Parascandolo, Parish, Parparita, Passos, Pavlov, Peng,
  Perelman, de~Avila Belbute~Peres, Petrov, de~Oliveira~Pinto, Pokorny,
  Pokrass, Pong, Powell, Power, Power, Proehl, Puri, Radford, Rae, Ramesh,
  Raymond, Real, Rimbach, Ross, Rotsted, Roussez, Ryder, Saltarelli, Sanders,
  Santurkar, Sastry, Schmidt, Schnurr, Schulman, Selsam, Sheppard, Sherbakov,
  Shieh, Shoker, Shyam, Sidor, Sigler, Simens, Sitkin, Slama, Sohl, Sokolowsky,
  Song, Staudacher, Such, Summers, Sutskever, Tang, Tezak, Thompson, Tillet,
  Tootoonchian, Tseng, Tuggle, Turley, Tworek, Uribe, Vallone, Vijayvergiya,
  Voss, Wainwright, Wang, Wang, Wang, Ward, Wei, Weinmann, Welihinda, Welinder,
  Weng, Weng, Wiethoff, Willner, Winter, Wolrich, Wong, Workman, Wu, Wu, Wu,
  Xiao, Xu, Yoo, Yu, Yuan, Zaremba, Zellers, Zhang, Zhang, Zhao, Zheng, Zhuang,
  Zhuk, and Zoph]{Achiam2023GPT4TR}
Achiam, O.~J., Adler, S., Agarwal, S., Ahmad, L., Akkaya, I., Aleman, F.~L.,
  Almeida, D., Altenschmidt, J., Altman, S., Anadkat, S., Avila, R.,
  Babuschkin, I., Balaji, S., Balcom, V., Baltescu, P., Bao, H., Bavarian, M.,
  Belgum, J., Bello, I., Berdine, J., Bernadett-Shapiro, G., Berner, C.,
  Bogdonoff, L., Boiko, O., Boyd, M., Brakman, A.-L., Brockman, G., Brooks, T.,
  Brundage, M., Button, K., Cai, T., Campbell, R., Cann, A., Carey, B.,
  Carlson, C., Carmichael, R., Chan, B., Chang, C., Chantzis, F., Chen, D.,
  Chen, S., Chen, R., Chen, J., Chen, M., Chess, B., Cho, C., Chu, C., Chung,
  H.~W., Cummings, D., Currier, J., Dai, Y., Decareaux, C., Degry, T., Deutsch,
  N., Deville, D., Dhar, A., Dohan, D., Dowling, S., Dunning, S., Ecoffet, A.,
  Eleti, A., Eloundou, T., Farhi, D., Fedus, L., Felix, N., Fishman, S.~P.,
  Forte, J., Fulford, I., Gao, L., Georges, E., Gibson, C., Goel, V., Gogineni,
  T., Goh, G., Gontijo-Lopes, R., Gordon, J., Grafstein, M., Gray, S., Greene,
  R., Gross, J., Gu, S.~S., Guo, Y., Hallacy, C., Han, J., Harris, J., He, Y.,
  Heaton, M., Heidecke, J., Hesse, C., Hickey, A., Hickey, W., Hoeschele, P.,
  Houghton, B., Hsu, K., Hu, S., Hu, X., Huizinga, J., Jain, S., Jain, S.,
  Jang, J., Jiang, A., Jiang, R., Jin, H., Jin, D., Jomoto, S., Jonn, B., Jun,
  H., Kaftan, T., Kaiser, L., Kamali, A., Kanitscheider, I., Keskar, N.~S.,
  Khan, T., Kilpatrick, L., Kim, J.~W., Kim, C., Kim, Y., Kirchner, H., Kiros,
  J.~R., Knight, M., Kokotajlo, D., Kondraciuk, L., Kondrich, A.,
  Konstantinidis, A., Kosic, K., Krueger, G., Kuo, V., Lampe, M., Lan, I., Lee,
  T., Leike, J., Leung, J., Levy, D., Li, C.~M., Lim, R., Lin, M., Lin, S.,
  Litwin, M., Lopez, T., Lowe, R., Lue, P., Makanju, A.~A., Malfacini, K.,
  Manning, S., Markov, T., Markovski, Y., Martin, B., Mayer, K., Mayne, A.,
  McGrew, B., McKinney, S.~M., McLeavey, C., McMillan, P., McNeil, J., Medina,
  D., Mehta, A., Menick, J., Metz, L., Mishchenko, A., Mishkin, P., Monaco, V.,
  Morikawa, E., Mossing, D.~P., Mu, T., Murati, M., Murk, O., M'ely, D., Nair,
  A., Nakano, R., Nayak, R., Neelakantan, A., Ngo, R., Noh, H., Long, O.,
  O'Keefe, C., Pachocki, J.~W., Paino, A., Palermo, J., Pantuliano, A.,
  Parascandolo, G., Parish, J., Parparita, E., Passos, A., Pavlov, M., Peng,
  A., Perelman, A., de~Avila Belbute~Peres, F., Petrov, M., de~Oliveira~Pinto,
  H.~P., Pokorny, M., Pokrass, M., Pong, V.~H., Powell, T., Power, A., Power,
  B., Proehl, E., Puri, R., Radford, A., Rae, J., Ramesh, A., Raymond, C.,
  Real, F., Rimbach, K., Ross, C., Rotsted, B., Roussez, H., Ryder, N.,
  Saltarelli, M.~D., Sanders, T., Santurkar, S., Sastry, G., Schmidt, H.,
  Schnurr, D., Schulman, J., Selsam, D., Sheppard, K., Sherbakov, T., Shieh,
  J., Shoker, S., Shyam, P., Sidor, S., Sigler, E., Simens, M., Sitkin, J.,
  Slama, K., Sohl, I., Sokolowsky, B.~D., Song, Y., Staudacher, N., Such,
  F.~P., Summers, N., Sutskever, I., Tang, J., Tezak, N.~A., Thompson, M.,
  Tillet, P., Tootoonchian, A., Tseng, E., Tuggle, P., Turley, N., Tworek, J.,
  Uribe, J. F.~C., Vallone, A., Vijayvergiya, A., Voss, C., Wainwright, C.,
  Wang, J.~J., Wang, A., Wang, B., Ward, J., Wei, J., Weinmann, C., Welihinda,
  A., Welinder, P., Weng, J., Weng, L., Wiethoff, M., Willner, D., Winter, C.,
  Wolrich, S., Wong, H., Workman, L., Wu, S., Wu, J., Wu, M., Xiao, K., Xu, T.,
  Yoo, S., Yu, K., Yuan, Q., Zaremba, W., Zellers, R., Zhang, C., Zhang, M.,
  Zhao, S., Zheng, T., Zhuang, J., Zhuk, W., and Zoph, B.
\newblock Gpt-4 technical report.
\newblock 2023.
\newblock URL \url{https://api.semanticscholar.org/CorpusID:257532815}.

\bibitem[Aher et~al.(2023)Aher, Arriaga, and Kalai]{aher2023using}
Aher, G.~V., Arriaga, R.~I., and Kalai, A.~T.
\newblock Using large language models to simulate multiple humans and replicate
  human subject studies.
\newblock In \emph{International Conference on Machine Learning}, pp.\
  337--371. PMLR, 2023.

\bibitem[Anthropic(2023)]{anthropic2023}
Anthropic.
\newblock Introducing claude, 2023.
\newblock URL \url{https://www.anthropic.com/index/introducing-claude}.

\bibitem[Argyle et~al.(2023)Argyle, Busby, Fulda, Gubler, Rytting, and
  Wingate]{out-of-one}
Argyle, L., Busby, E., Fulda, N., Gubler, J., Rytting, C., and Wingate, D.
\newblock Out of one, many: Using language models to simulate human samples.
\newblock \emph{Political Analysis}, 31:\penalty0 1--15, 02 2023.
\newblock \doi{10.1017/pan.2023.2}.

\bibitem[Arnesen \& Peters(2018)Arnesen and
  Peters]{doi:10.1177/0010414017720702}
Arnesen, S. and Peters, Y.
\newblock The legitimacy of representation: How descriptive, formal, and
  responsiveness representation affect the acceptability of political
  decisions.
\newblock \emph{Comparative Political Studies}, 51\penalty0 (7):\penalty0
  868--899, 2018.
\newblock \doi{10.1177/0010414017720702}.
\newblock URL \url{https://doi.org/10.1177/0010414017720702}.

\bibitem[Aroyo et~al.(2023)Aroyo, Taylor, Diaz, Homan, Parrish, Serapio-Garcia,
  Prabhakaran, and Wang]{aroyo2023dices}
Aroyo, L., Taylor, A.~S., Diaz, M., Homan, C.~M., Parrish, A., Serapio-Garcia,
  G., Prabhakaran, V., and Wang, D.
\newblock Dices dataset: Diversity in conversational ai evaluation for safety,
  2023.

\bibitem[Askell et~al.(2021)Askell, Bai, Chen, Drain, Ganguli, Henighan, Jones,
  Joseph, Mann, DasSarma, Elhage, Hatfield-Dodds, Hernandez, Kernion, Ndousse,
  Olsson, Amodei, Brown, Clark, McCandlish, Olah, and
  Kaplan]{askell2021general}
Askell, A., Bai, Y., Chen, A., Drain, D., Ganguli, D., Henighan, T., Jones, A.,
  Joseph, N., Mann, B., DasSarma, N., Elhage, N., Hatfield-Dodds, Z.,
  Hernandez, D., Kernion, J., Ndousse, K., Olsson, C., Amodei, D., Brown, T.,
  Clark, J., McCandlish, S., Olah, C., and Kaplan, J.
\newblock A general language assistant as a laboratory for alignment, 2021.

\bibitem[Bai et~al.(2022{\natexlab{a}})Bai, Jones, Ndousse, Askell, Chen,
  DasSarma, Drain, Fort, Ganguli, Henighan, Joseph, Kadavath, Kernion, Conerly,
  El-Showk, Elhage, Hatfield-Dodds, Hernandez, Hume, Johnston, Kravec, Lovitt,
  Nanda, Olsson, Amodei, Brown, Clark, McCandlish, Olah, Mann, and
  Kaplan]{bai2022training}
Bai, Y., Jones, A., Ndousse, K., Askell, A., Chen, A., DasSarma, N., Drain, D.,
  Fort, S., Ganguli, D., Henighan, T., Joseph, N., Kadavath, S., Kernion, J.,
  Conerly, T., El-Showk, S., Elhage, N., Hatfield-Dodds, Z., Hernandez, D.,
  Hume, T., Johnston, S., Kravec, S., Lovitt, L., Nanda, N., Olsson, C.,
  Amodei, D., Brown, T., Clark, J., McCandlish, S., Olah, C., Mann, B., and
  Kaplan, J.
\newblock Training a helpful and harmless assistant with reinforcement learning
  from human feedback, 2022{\natexlab{a}}.

\bibitem[Bai et~al.(2022{\natexlab{b}})Bai, Kadavath, Kundu, Askell, Kernion,
  Jones, Chen, Goldie, Mirhoseini, McKinnon, Chen, Olsson, Olah, Hernandez,
  Drain, Ganguli, Li, Tran-Johnson, Perez, Kerr, Mueller, Ladish, Landau,
  Ndousse, Lukosuite, Lovitt, Sellitto, Elhage, Schiefer, Mercado, DasSarma,
  Lasenby, Larson, Ringer, Johnston, Kravec, Showk, Fort, Lanham,
  Telleen-Lawton, Conerly, Henighan, Hume, Bowman, Hatfield-Dodds, Mann,
  Amodei, Joseph, McCandlish, Brown, and Kaplan]{bai2022constitutional}
Bai, Y., Kadavath, S., Kundu, S., Askell, A., Kernion, J., Jones, A., Chen, A.,
  Goldie, A., Mirhoseini, A., McKinnon, C., Chen, C., Olsson, C., Olah, C.,
  Hernandez, D., Drain, D., Ganguli, D., Li, D., Tran-Johnson, E., Perez, E.,
  Kerr, J., Mueller, J., Ladish, J., Landau, J., Ndousse, K., Lukosuite, K.,
  Lovitt, L., Sellitto, M., Elhage, N., Schiefer, N., Mercado, N., DasSarma,
  N., Lasenby, R., Larson, R., Ringer, S., Johnston, S., Kravec, S., Showk,
  S.~E., Fort, S., Lanham, T., Telleen-Lawton, T., Conerly, T., Henighan, T.,
  Hume, T., Bowman, S.~R., Hatfield-Dodds, Z., Mann, B., Amodei, D., Joseph,
  N., McCandlish, S., Brown, T., and Kaplan, J.
\newblock Constitutional ai: Harmlessness from ai feedback, 2022{\natexlab{b}}.

\bibitem[Bakker et~al.(2022)Bakker, Chadwick, Sheahan, Tessler,
  Campbell-Gillingham, Balaguer, McAleese, Glaese, Aslanides, Botvinick, and
  Summerfield]{bakker2022finetuning}
Bakker, M.~A., Chadwick, M.~J., Sheahan, H.~R., Tessler, M.~H.,
  Campbell-Gillingham, L., Balaguer, J., McAleese, N., Glaese, A., Aslanides,
  J., Botvinick, M.~M., and Summerfield, C.
\newblock Fine-tuning language models to find agreement among humans with
  diverse preferences, 2022.

\bibitem[Berlin(1969)]{berlin1969two}
Berlin, I.
\newblock Two concepts of liberty.
\newblock In \emph{Four Essays on Liberty}, pp.\  118–172. Oxford University
  Press, Oxford, 1969.

\bibitem[Bobu et~al.(2023)Bobu, Peng, Agrawal, Shah, and
  Dragan]{bobu2023aligning}
Bobu, A., Peng, A., Agrawal, P., Shah, J., and Dragan, A.~D.
\newblock Aligning robot and human representations.
\newblock \emph{arXiv preprint arXiv:2302.01928}, 2023.

\bibitem[Bommasani et~al.(2021)Bommasani, Hudson, Adeli, Altman, Arora, von
  Arx, Bernstein, Bohg, Bosselut, Brunskill, Brynjolfsson, Buch, Card,
  Castellon, Chatterji, Chen, Creel, Davis, Demszky, Donahue, Doumbouya,
  Durmus, Ermon, Etchemendy, Ethayarajh, Fei-Fei, Finn, Gale, Gillespie, Goel,
  Goodman, Grossman, Guha, Hashimoto, Henderson, Hewitt, Ho, Hong, Hsu, Huang,
  Icard, Jain, Jurafsky, Kalluri, Karamcheti, Keeling, Khani, Khattab, Koh,
  Krass, Krishna, Kuditipudi, Kumar, Ladhak, Lee, Lee, Leskovec, Levent, Li,
  Li, Ma, Malik, Manning, Mirchandani, Mitchell, Munyikwa, Nair, Narayan,
  Narayanan, Newman, Nie, Niebles, Nilforoshan, Nyarko, Ogut, Orr,
  Papadimitriou, Park, Piech, Portelance, Potts, Raghunathan, Reich, Ren, Rong,
  Roohani, Ruiz, Ryan, R'e, Sadigh, Sagawa, Santhanam, Shih, Srinivasan,
  Tamkin, Taori, Thomas, Tram{\`e}r, Wang, Wang, Wu, Wu, Wu, Xie, Yasunaga,
  You, Zaharia, Zhang, Zhang, Zhang, Zhang, Zheng, Zhou, and
  Liang]{Bommasani2021OnTO}
Bommasani, R., Hudson, D.~A., Adeli, E., Altman, R., Arora, S., von Arx, S.,
  Bernstein, M.~S., Bohg, J., Bosselut, A., Brunskill, E., Brynjolfsson, E.,
  Buch, S., Card, D., Castellon, R., Chatterji, N.~S., Chen, A.~S., Creel,
  K.~A., Davis, J., Demszky, D., Donahue, C., Doumbouya, M., Durmus, E., Ermon,
  S., Etchemendy, J., Ethayarajh, K., Fei-Fei, L., Finn, C., Gale, T.,
  Gillespie, L., Goel, K., Goodman, N.~D., Grossman, S., Guha, N., Hashimoto,
  T., Henderson, P., Hewitt, J., Ho, D.~E., Hong, J., Hsu, K., Huang, J.,
  Icard, T.~F., Jain, S., Jurafsky, D., Kalluri, P., Karamcheti, S., Keeling,
  G., Khani, F., Khattab, O., Koh, P.~W., Krass, M.~S., Krishna, R.,
  Kuditipudi, R., Kumar, A., Ladhak, F., Lee, M., Lee, T., Leskovec, J.,
  Levent, I., Li, X.~L., Li, X., Ma, T., Malik, A., Manning, C.~D.,
  Mirchandani, S., Mitchell, E., Munyikwa, Z., Nair, S., Narayan, A.,
  Narayanan, D., Newman, B., Nie, A., Niebles, J.~C., Nilforoshan, H., Nyarko,
  J.~F., Ogut, G., Orr, L.~J., Papadimitriou, I., Park, J.~S., Piech, C.,
  Portelance, E., Potts, C., Raghunathan, A., Reich, R., Ren, H., Rong, F.,
  Roohani, Y.~H., Ruiz, C., Ryan, J., R'e, C., Sadigh, D., Sagawa, S.,
  Santhanam, K., Shih, A., Srinivasan, K.~P., Tamkin, A., Taori, R., Thomas,
  A.~W., Tram{\`e}r, F., Wang, R.~E., Wang, W., Wu, B., Wu, J., Wu, Y., Xie,
  S.~M., Yasunaga, M., You, J., Zaharia, M.~A., Zhang, M., Zhang, T., Zhang,
  X., Zhang, Y., Zheng, L., Zhou, K., and Liang, P.
\newblock On the opportunities and risks of foundation models.
\newblock \emph{ArXiv}, abs/2108.07258, 2021.
\newblock URL \url{https://arxiv.org/pdf/2108.07258.pdf}.

\bibitem[Bowman et~al.(2022)Bowman, Hyun, Perez, Chen, Pettit, Heiner,
  Lukošiūtė, Askell, Jones, Chen, Goldie, Mirhoseini, McKinnon, Olah,
  Amodei, Amodei, Drain, Li, Tran-Johnson, Kernion, Kerr, Mueller, Ladish,
  Landau, Ndousse, Lovitt, Elhage, Schiefer, Joseph, Mercado, DasSarma, Larson,
  McCandlish, Kundu, Johnston, Kravec, Showk, Fort, Telleen-Lawton, Brown,
  Henighan, Hume, Bai, Hatfield-Dodds, Mann, and Kaplan]{bowman2022measuring}
Bowman, S.~R., Hyun, J., Perez, E., Chen, E., Pettit, C., Heiner, S.,
  Lukošiūtė, K., Askell, A., Jones, A., Chen, A., Goldie, A., Mirhoseini,
  A., McKinnon, C., Olah, C., Amodei, D., Amodei, D., Drain, D., Li, D.,
  Tran-Johnson, E., Kernion, J., Kerr, J., Mueller, J., Ladish, J., Landau, J.,
  Ndousse, K., Lovitt, L., Elhage, N., Schiefer, N., Joseph, N., Mercado, N.,
  DasSarma, N., Larson, R., McCandlish, S., Kundu, S., Johnston, S., Kravec,
  S., Showk, S.~E., Fort, S., Telleen-Lawton, T., Brown, T., Henighan, T.,
  Hume, T., Bai, Y., Hatfield-Dodds, Z., Mann, B., and Kaplan, J.
\newblock Measuring progress on scalable oversight for large language models,
  2022.

\bibitem[Boykoff \& Boykoff(2004)Boykoff and Boykoff]{BOYKOFF2004125}
Boykoff, M.~T. and Boykoff, J.~M.
\newblock Balance as bias: global warming and the us prestige press.
\newblock \emph{Global Environmental Change}, 14\penalty0 (2):\penalty0
  125--136, 2004.
\newblock ISSN 0959-3780.
\newblock \doi{https://doi.org/10.1016/j.gloenvcha.2003.10.001}.
\newblock URL
  \url{https://www.sciencedirect.com/science/article/pii/S0959378003000669}.

\bibitem[Brown et~al.(2020)Brown, Mann, Ryder, Subbiah, Kaplan, Dhariwal,
  Neelakantan, Shyam, Sastry, Askell, Agarwal, Herbert-Voss, Krueger, Henighan,
  Child, Ramesh, Ziegler, Wu, Winter, Hesse, Chen, Sigler, Litwin, Gray, Chess,
  Clark, Berner, McCandlish, Radford, Sutskever, and
  Amodei]{Brown2020LanguageMA}
Brown, T.~B., Mann, B., Ryder, N., Subbiah, M., Kaplan, J., Dhariwal, P.,
  Neelakantan, A., Shyam, P., Sastry, G., Askell, A., Agarwal, S.,
  Herbert-Voss, A., Krueger, G., Henighan, T.~J., Child, R., Ramesh, A.,
  Ziegler, D.~M., Wu, J., Winter, C., Hesse, C., Chen, M., Sigler, E., Litwin,
  M., Gray, S., Chess, B., Clark, J., Berner, C., McCandlish, S., Radford, A.,
  Sutskever, I., and Amodei, D.
\newblock Language models are few-shot learners.
\newblock \emph{ArXiv}, abs/2005.14165, 2020.
\newblock URL \url{https://api.semanticscholar.org/CorpusID:218971783}.

\bibitem[Buttrick(2024)]{Buttrick2024}
Buttrick, N.
\newblock Studying large language models as compression algorithms for human
  culture.
\newblock \emph{Trends in Cognitive Sciences}, S1364-6613\penalty0
  (24):\penalty0 00001--9, 2024.
\newblock \doi{10.1016/j.tics.2024.01.001}.
\newblock Epub ahead of print.

\bibitem[Casper et~al.(2023)Casper, Davies, Shi, Gilbert, Scheurer, Rando,
  Freedman, Korbak, Lindner, Freire, Wang, Marks, S{\'e}gerie, Carroll, Peng,
  Christoffersen, Damani, Slocum, Anwar, Siththaranjan, Nadeau, Michaud, Pfau,
  Krasheninnikov, Chen, di~Langosco, Hase, Biyik, Dragan, Krueger, Sadigh, and
  Hadfield-Menell]{Casper2023OpenPA}
Casper, S., Davies, X., Shi, C., Gilbert, T.~K., Scheurer, J., Rando, J.,
  Freedman, R., Korbak, T., Lindner, D., Freire, P., Wang, T., Marks, S.,
  S{\'e}gerie, C.-R., Carroll, M., Peng, A., Christoffersen, P.~J., Damani, M.,
  Slocum, S., Anwar, U., Siththaranjan, A., Nadeau, M., Michaud, E.~J., Pfau,
  J., Krasheninnikov, D., Chen, X., di~Langosco, L.~L., Hase, P., Biyik, E.,
  Dragan, A.~D., Krueger, D., Sadigh, D., and Hadfield-Menell, D.
\newblock Open problems and fundamental limitations of reinforcement learning
  from human feedback.
\newblock \emph{ArXiv}, abs/2307.15217, 2023.
\newblock URL \url{https://api.semanticscholar.org/CorpusID:260316010}.

\bibitem[Chen et~al.(2023)Chen, Liu, Huang, Wu, Liu, Jiang, Pu, Lei, Chen,
  Wang, Lian, and Chen]{chen2023large}
Chen, J., Liu, Z., Huang, X., Wu, C., Liu, Q., Jiang, G., Pu, Y., Lei, Y.,
  Chen, X., Wang, X., Lian, D., and Chen, E.
\newblock When large language models meet personalization: Perspectives of
  challenges and opportunities, 2023.

\bibitem[Chen et~al.(2021)Chen, Lu, Rajeswaran, Lee, Grover, Laskin, Abbeel,
  Srinivas, and Mordatch]{chen2021decision}
Chen, L., Lu, K., Rajeswaran, A., Lee, K., Grover, A., Laskin, M., Abbeel, P.,
  Srinivas, A., and Mordatch, I.
\newblock Decision transformer: Reinforcement learning via sequence modeling,
  2021.

\bibitem[Cotra(2021)]{cotra2021cold}
Cotra, A.
\newblock Why ai alignment could be hard with modern deep learning.
\newblock
  \url{https://www.cold-takes.com/why-ai-alignment-could-be-hard-with-modern-deep-learning/},
  2021.

\bibitem[Crenshaw(1989)]{crenshaw1989demarginalizing}
Crenshaw, K.
\newblock Demarginalizing the intersection of race and sex: A black feminist
  critique of antidiscrimination doctrine, feminist theory and antiracist
  politics.
\newblock \emph{The University of Chicago Legal Forum}, 140:\penalty0 139--167,
  1989.

\bibitem[Danry et~al.(2023)Danry, Pataranutaporn, Mao, and
  Maes]{dontJustTellMe}
Danry, V., Pataranutaporn, P., Mao, Y., and Maes, P.
\newblock Don’t just tell me, ask me: Ai systems that intelligently frame
  explanations as questions improve human logical discernment accuracy over
  causal ai explanations.
\newblock In \emph{Proceedings of the 2023 CHI Conference on Human Factors in
  Computing Systems}, CHI '23, New York, NY, USA, 2023. Association for
  Computing Machinery.
\newblock ISBN 9781450394215.
\newblock \doi{10.1145/3544548.3580672}.
\newblock URL \url{https://doi.org/10.1145/3544548.3580672}.

\bibitem[de~Tocqueville(1835)]{tocqueville1835democracy}
de~Tocqueville, A.
\newblock \emph{Democracy in America}.
\newblock 1835.

\bibitem[Durmus et~al.(2023)Durmus, Nyugen, Liao, Schiefer, Askell, Bakhtin,
  Chen, Hatfield-Dodds, Hernandez, Joseph, Lovitt, McCandlish, Sikder, Tamkin,
  Thamkul, Kaplan, Clark, and Ganguli]{durmus2023measuring}
Durmus, E., Nyugen, K., Liao, T.~I., Schiefer, N., Askell, A., Bakhtin, A.,
  Chen, C., Hatfield-Dodds, Z., Hernandez, D., Joseph, N., Lovitt, L.,
  McCandlish, S., Sikder, O., Tamkin, A., Thamkul, J., Kaplan, J., Clark, J.,
  and Ganguli, D.
\newblock Towards measuring the representation of subjective global opinions in
  language models, 2023.
\newblock URL \url{https://api.semanticscholar.org/CorpusID:259275051}.

\bibitem[Ethayarajh \& Jurafsky(2020)Ethayarajh and
  Jurafsky]{Ethayarajh2020UtilityII}
Ethayarajh, K. and Jurafsky, D.
\newblock Utility is in the eye of the user: A critique of nlp leaderboard
  design.
\newblock In \emph{Conference on Empirical Methods in Natural Language
  Processing}, 2020.
\newblock URL \url{https://api.semanticscholar.org/CorpusID:235408131}.

\bibitem[Ethayarajh \& Jurafsky(2022)Ethayarajh and
  Jurafsky]{ethayarajh-jurafsky-2022-authenticity}
Ethayarajh, K. and Jurafsky, D.
\newblock The authenticity gap in human evaluation.
\newblock In Goldberg, Y., Kozareva, Z., and Zhang, Y. (eds.),
  \emph{Proceedings of the 2022 Conference on Empirical Methods in Natural
  Language Processing}, pp.\  6056--6070, Abu Dhabi, United Arab Emirates,
  December 2022. Association for Computational Linguistics.
\newblock \doi{10.18653/v1/2022.emnlp-main.406}.
\newblock URL \url{https://aclanthology.org/2022.emnlp-main.406}.

\bibitem[Feng et~al.(2023)Feng, Park, Liu, and Tsvetkov]{feng2023pretraining}
Feng, S., Park, C.~Y., Liu, Y., and Tsvetkov, Y.
\newblock From pretraining data to language models to downstream tasks:
  Tracking the trails of political biases leading to unfair nlp models, 2023.

\bibitem[Flanigan et~al.(2021)Flanigan, G{\"o}lz, Gupta, Hennig, and
  Procaccia]{citizens-assemblies}
Flanigan, B., G{\"o}lz, P., Gupta, A., Hennig, B., and Procaccia, A.~D.
\newblock Fair algorithms for selecting citizens'assemblies.
\newblock \emph{Nature}, 596\penalty0 (7873):\penalty0 548--552, 2021.
\newblock \doi{10.1038/s41586-021-03788-6}.
\newblock URL \url{https://doi.org/10.1038/s41586-021-03788-6}.

\bibitem[Fleisig et~al.(2023)Fleisig, Abebe, and Klein]{fleisig_when_2023}
Fleisig, E., Abebe, R., and Klein, D.
\newblock When the {Majority} is {Wrong}: {Modeling} {Annotator} {Disagreement}
  for {Subjective} {Tasks}, November 2023.
\newblock URL \url{http://arxiv.org/abs/2305.06626}.
\newblock arXiv:2305.06626 [cs].

\bibitem[Gabriel(2020)]{gabriel2020}
Gabriel, I.
\newblock Artificial intelligence, values, and alignment.
\newblock \emph{Minds and Machines}, 30\penalty0 (3):\penalty0 411--437, 2020.
\newblock \doi{10.1007/s11023-020-09539-2}.
\newblock URL \url{https://doi.org/10.1007/s11023-020-09539-2}.

\bibitem[Girotra et~al.(2023)Girotra, Meincke, Terwiesch, and
  Ulrich]{Girotra2023}
Girotra, K., Meincke, L., Terwiesch, C., and Ulrich, K.~T.
\newblock Ideas are dimes a dozen: Large language models for idea generation in
  innovation.
\newblock \url{https://ssrn.com/abstract=4526071}, July 2023.
\newblock Available at SSRN: https://ssrn.com/abstract=4526071 or
  http://dx.doi.org/10.2139/ssrn.4526071.

\bibitem[Glaese et~al.(2022)Glaese, McAleese, Trębacz, Aslanides, Firoiu,
  Ewalds, Rauh, Weidinger, Chadwick, Thacker, Campbell-Gillingham, Uesato,
  Huang, Comanescu, Yang, See, Dathathri, Greig, Chen, Fritz, Elias, Green,
  Mokrá, Fernando, Wu, Foley, Young, Gabriel, Isaac, Mellor, Hassabis,
  Kavukcuoglu, Hendricks, and Irving]{glaese2022improving}
Glaese, A., McAleese, N., Trębacz, M., Aslanides, J., Firoiu, V., Ewalds, T.,
  Rauh, M., Weidinger, L., Chadwick, M., Thacker, P., Campbell-Gillingham, L.,
  Uesato, J., Huang, P.-S., Comanescu, R., Yang, F., See, A., Dathathri, S.,
  Greig, R., Chen, C., Fritz, D., Elias, J.~S., Green, R., Mokrá, S.,
  Fernando, N., Wu, B., Foley, R., Young, S., Gabriel, I., Isaac, W., Mellor,
  J., Hassabis, D., Kavukcuoglu, K., Hendricks, L.~A., and Irving, G.
\newblock Improving alignment of dialogue agents via targeted human judgements,
  2022.

\bibitem[Gordon et~al.(2022)Gordon, Lam, Park, Patel, Hancock, Hashimoto, and
  Bernstein]{Gordon_2022}
Gordon, M.~L., Lam, M.~S., Park, J.~S., Patel, K., Hancock, J., Hashimoto, T.,
  and Bernstein, M.~S.
\newblock Jury learning: Integrating dissenting voices into machine learning
  models.
\newblock In \emph{CHI Conference on Human Factors in Computing Systems}, CHI
  ’22. ACM, April 2022.
\newblock \doi{10.1145/3491102.3502004}.
\newblock URL \url{http://dx.doi.org/10.1145/3491102.3502004}.

\bibitem[Guerreiro et~al.(2020)Guerreiro, Fonseca, and
  Paquete]{Guerreiro2020TheHI}
Guerreiro, A.~P., Fonseca, C.~M., and Paquete, L.
\newblock The hypervolume indicator.
\newblock \emph{ACM Computing Surveys (CSUR)}, 54:\penalty0 1--42, 2020.
\newblock URL \url{https://api.semanticscholar.org/CorpusID:218470181}.

\bibitem[Haraway(1988)]{situatedKnowledges}
Haraway, D.
\newblock Situated knowledges: The science question in feminism and the
  privilege of partial perspective.
\newblock \emph{Feminist Studies}, 14\penalty0 (3):\penalty0 575--599, 1988.
\newblock ISSN 00463663.
\newblock URL \url{http://www.jstor.org/stable/3178066}.

\bibitem[Harsanyi et~al.(1988)Harsanyi, Selten, et~al.]{harsanyi1988general}
Harsanyi, J.~C., Selten, R., et~al.
\newblock A general theory of equilibrium selection in games.
\newblock \emph{MIT Press Books}, 1, 1988.

\bibitem[Hartmann et~al.(2023)Hartmann, Schwenzow, and
  Witte]{hartmann2023political}
Hartmann, J., Schwenzow, J., and Witte, M.
\newblock The political ideology of conversational ai: Converging evidence on
  chatgpt's pro-environmental, left-libertarian orientation, 2023.

\bibitem[Hayati et~al.(2023)Hayati, Lee, Rajagopal, and Kang]{Hayati2023HowFC}
Hayati, S.~A., Lee, M., Rajagopal, D., and Kang, D.
\newblock How far can we extract diverse perspectives from large language
  models? criteria-based diversity prompting!
\newblock \emph{ArXiv}, abs/2311.09799, 2023.
\newblock URL \url{https://api.semanticscholar.org/CorpusID:265220883}.

\bibitem[Hayes et~al.(2022)Hayes, Rădulescu, Bargiacchi, Källström,
  Macfarlane, Reymond, Verstraeten, Zintgraf, Dazeley, Heintz, Howley,
  Irissappane, Mannion, Nowé, Ramos, Restelli, Vamplew, and
  Roijers]{Hayes_2022}
Hayes, C.~F., Rădulescu, R., Bargiacchi, E., Källström, J., Macfarlane, M.,
  Reymond, M., Verstraeten, T., Zintgraf, L.~M., Dazeley, R., Heintz, F.,
  Howley, E., Irissappane, A.~A., Mannion, P., Nowé, A., Ramos, G., Restelli,
  M., Vamplew, P., and Roijers, D.~M.
\newblock A practical guide to multi-objective reinforcement learning and
  planning.
\newblock \emph{Autonomous Agents and Multi-Agent Systems}, 36\penalty0
  (1):\penalty0 26, April 2022.
\newblock ISSN 1573-7454.
\newblock \doi{10.1007/s10458-022-09552-y}.
\newblock URL \url{http://dx.doi.org/10.1007/s10458-022-09552-y}.

\bibitem[Hendrycks et~al.(2020)Hendrycks, Burns, Basart, Critch, Li, Song, and
  Steinhardt]{hendrycks2020aligning}
Hendrycks, D., Burns, C., Basart, S., Critch, A., Li, J., Song, D., and
  Steinhardt, J.
\newblock Aligning ai with shared human values.
\newblock \emph{arXiv preprint arXiv:2008.02275}, 2020.

\bibitem[Hendrycks et~al.(2023)Hendrycks, Burns, Basart, Critch, Li, Song, and
  Steinhardt]{hendrycks2023aligning}
Hendrycks, D., Burns, C., Basart, S., Critch, A., Li, J., Song, D., and
  Steinhardt, J.
\newblock Aligning ai with shared human values, 2023.

\bibitem[Henrich et~al.(2010)Henrich, Heine, and Norenzayan]{weird}
Henrich, J., Heine, S.~J., and Norenzayan, A.
\newblock The weirdest people in the world?
\newblock \emph{Behavioral and Brain Sciences}, 33\penalty0 (2-3):\penalty0
  61--83, 2010.
\newblock URL \url{http://www2.psych.ubc.ca/~henrich/audiofiles/WEIRD1.mp3}.

\bibitem[Hosking et~al.(2023)Hosking, Blunsom, and Bartolo]{Hosking2023HumanFI}
Hosking, T., Blunsom, P., and Bartolo, M.
\newblock Human feedback is not gold standard.
\newblock \emph{ArXiv}, abs/2309.16349, 2023.
\newblock URL \url{https://api.semanticscholar.org/CorpusID:263134280}.

\bibitem[Hsieh \& Andersson(2021)Hsieh and
  Andersson]{sep-value-incommensurable}
Hsieh, N.-h. and Andersson, H.
\newblock {Incommensurable Values}.
\newblock In Zalta, E.~N. (ed.), \emph{The {Stanford} Encyclopedia of
  Philosophy}. Metaphysics Research Lab, Stanford University, {F}all 2021
  edition, 2021.

\bibitem[Hwang et~al.(2023)Hwang, Majumder, and Tandon]{hwang_aligning_2023}
Hwang, E., Majumder, B.~P., and Tandon, N.
\newblock Aligning {Language} {Models} to {User} {Opinions}.
\newblock 2023.
\newblock \doi{10.48550/ARXIV.2305.14929}.
\newblock URL \url{https://arxiv.org/abs/2305.14929}.
\newblock Publisher: arXiv Version Number: 1.

\bibitem[Imundo \& Rapp(2021)Imundo and Rapp]{falseBalance}
Imundo, M. and Rapp, D.
\newblock When fairness is flawed: Effects of false balance reporting and
  weight-of-evidence statements on beliefs and perceptions of climate change.
\newblock \emph{Journal of Applied Research in Memory and Cognition}, 11, 10
  2021.
\newblock \doi{10.1016/j.jarmac.2021.10.002}.

\bibitem[Jakesch et~al.(2023)Jakesch, Bhat, Buschek, Zalmanson, and
  Naaman]{10.1145/3544548.3581196}
Jakesch, M., Bhat, A., Buschek, D., Zalmanson, L., and Naaman, M.
\newblock Co-writing with opinionated language models affects users’ views.
\newblock In \emph{Proceedings of the 2023 CHI Conference on Human Factors in
  Computing Systems}, CHI '23, New York, NY, USA, 2023. Association for
  Computing Machinery.
\newblock ISBN 9781450394215.
\newblock \doi{10.1145/3544548.3581196}.
\newblock URL \url{https://doi.org/10.1145/3544548.3581196}.

\bibitem[Jang et~al.(2023)Jang, Kim, Lin, Wang, Hessel, Zettlemoyer,
  Hajishirzi, Choi, and Ammanabrolu]{jang2023personalized}
Jang, J., Kim, S., Lin, B.~Y., Wang, Y., Hessel, J., Zettlemoyer, L.,
  Hajishirzi, H., Choi, Y., and Ammanabrolu, P.
\newblock Personalized soups: Personalized large language model alignment via
  post-hoc parameter merging, 2023.

\bibitem[Ji et~al.(2024)Ji, Qiu, Chen, Zhang, Lou, Wang, Duan, He, Zhou, Zhang,
  Zeng, Ng, Dai, Pan, O'Gara, Lei, Xu, Tse, Fu, McAleer, Yang, Wang, Zhu, Guo,
  and Gao]{ji2024ai}
Ji, J., Qiu, T., Chen, B., Zhang, B., Lou, H., Wang, K., Duan, Y., He, Z.,
  Zhou, J., Zhang, Z., Zeng, F., Ng, K.~Y., Dai, J., Pan, X., O'Gara, A., Lei,
  Y., Xu, H., Tse, B., Fu, J., McAleer, S., Yang, Y., Wang, Y., Zhu, S.-C.,
  Guo, Y., and Gao, W.
\newblock Ai alignment: A comprehensive survey, 2024.

\bibitem[Ji et~al.(2021)Ji, Li, and Telgarsky]{ji2021earlystopped}
Ji, Z., Li, J.~D., and Telgarsky, M.
\newblock Early-stopped neural networks are consistent, 2021.

\bibitem[Jiang et~al.(2023)Jiang, Xu, Zhu, Han, Zhang, and
  Zhu]{jiang2023evaluating}
Jiang, G., Xu, M., Zhu, S.-C., Han, W., Zhang, C., and Zhu, Y.
\newblock Evaluating and inducing personality in pre-trained language models,
  2023.
\newblock URL \url{https://api.semanticscholar.org/CorpusID:258865158}.

\bibitem[Jiang et~al.(2022)Jiang, Beeferman, Roy, and
  Roy]{jiang2022communitylm}
Jiang, H., Beeferman, D., Roy, B., and Roy, D.
\newblock Communitylm: Probing partisan worldviews from language models, 2022.

\bibitem[Jung et~al.(2022)Jung, Qin, Welleck, Brahman, Bhagavatula, Bras, and
  Choi]{jung2022maieutic}
Jung, J., Qin, L., Welleck, S., Brahman, F., Bhagavatula, C., Bras, R.~L., and
  Choi, Y.
\newblock Maieutic prompting: Logically consistent reasoning with recursive
  explanations, 2022.

\bibitem[Kant(1788)]{critiquepracticalreason}
Kant, I.
\newblock \emph{Kant: Critique of Practical Reason}.
\newblock Cambridge Texts in the History of Philosophy. Cambridge University
  Press, 2 edition, 1788.
\newblock \doi{10.1017/CBO9781316136478}.

\bibitem[Kasirzadeh \& Gabriel(2022)Kasirzadeh and
  Gabriel]{kasirzadeh2022conversation}
Kasirzadeh, A. and Gabriel, I.
\newblock In conversation with artificial intelligence: aligning language
  models with human values, 2022.

\bibitem[Kekes(1993)]{kekes1993morality}
Kekes, J.
\newblock \emph{The Morality of Pluralism}.
\newblock Princeton University Press, Princeton, 1993.

\bibitem[Kim \& Lee(2023)Kim and Lee]{kim2023ai}
Kim, J. and Lee, B.
\newblock Ai-augmented surveys: Leveraging large language models for opinion
  prediction in nationally representative surveys.
\newblock \emph{arXiv preprint arXiv:2305.09620}, 2023.

\bibitem[Kirk et~al.(2024)Kirk, Mediratta, Nalmpantis, Luketina, Hambro,
  Grefenstette, and Raileanu]{kirk2024understanding}
Kirk, R., Mediratta, I., Nalmpantis, C., Luketina, J., Hambro, E.,
  Grefenstette, E., and Raileanu, R.
\newblock Understanding the effects of rlhf on llm generalisation and
  diversity, 2024.

\bibitem[Koster et~al.(2022)Koster, Balaguer, Tacchetti, Weinstein, Zhu,
  Hauser, Williams, Campbell-Gillingham, Thacker, Botvinick, and
  Summerfield]{democratic}
Koster, R., Balaguer, J., Tacchetti, A., Weinstein, A., Zhu, T., Hauser, O.,
  Williams, D., Campbell-Gillingham, L., Thacker, P., Botvinick, M., and
  Summerfield, C.
\newblock Human-centred mechanism design with democratic ai.
\newblock \emph{Nature Human Behaviour}, 6\penalty0 (10):\penalty0 1398--1407,
  2022.
\newblock \doi{10.1038/s41562-022-01383-x}.
\newblock URL \url{https://doi.org/10.1038/s41562-022-01383-x}.

\bibitem[Kr{\"u}gel et~al.(2023)Kr{\"u}gel, Ostermaier, and
  Uhl]{inconsistent-moral}
Kr{\"u}gel, S., Ostermaier, A., and Uhl, M.
\newblock Chatgpt's inconsistent moral advice influences users'judgment.
\newblock \emph{Scientific Reports}, 13\penalty0 (1):\penalty0 4569, Apr 2023.
\newblock ISSN 2045-2322.
\newblock \doi{10.1038/s41598-023-31341-0}.
\newblock URL \url{https://doi.org/10.1038/s41598-023-31341-0}.

\bibitem[Landemore \& Page(2015)Landemore and Page]{landemore2015deliberation}
Landemore, H. and Page, S.~E.
\newblock Deliberation and disagreement: Problem solving, prediction, and
  positive dissensus.
\newblock \emph{Politics, philosophy \& economics}, 14\penalty0 (3):\penalty0
  229--254, 2015.

\bibitem[Leike et~al.(2018)Leike, Krueger, Everitt, Martic, Maini, and
  Legg]{leike2018scalable}
Leike, J., Krueger, D., Everitt, T., Martic, M., Maini, V., and Legg, S.
\newblock Scalable agent alignment via reward modeling: a research direction,
  2018.

\bibitem[Li et~al.(2023{\natexlab{a}})Li, Zhang, Mei, Wang, Hombaiah, Liang,
  and Bendersky]{li2023teach}
Li, C., Zhang, M., Mei, Q., Wang, Y., Hombaiah, S.~A., Liang, Y., and
  Bendersky, M.
\newblock Teach llms to personalize -- an approach inspired by writing
  education, 2023{\natexlab{a}}.

\bibitem[Li et~al.(2023{\natexlab{b}})Li, Mehrabi, Peris, Goyal, Chang,
  Galstyan, Zemel, and Gupta]{li_steerability_2023}
Li, J., Mehrabi, N., Peris, C., Goyal, P., Chang, K.-W., Galstyan, A., Zemel,
  R., and Gupta, R.
\newblock On the steerability of large language models toward data-driven
  personas.
\newblock 2023{\natexlab{b}}.
\newblock \doi{10.48550/ARXIV.2311.04978}.
\newblock URL \url{https://arxiv.org/abs/2311.04978}.
\newblock Publisher: arXiv Version Number: 1.

\bibitem[Liang et~al.(2023)Liang, Bommasani, Lee, Tsipras, Soylu, Yasunaga,
  Zhang, Narayanan, Wu, Kumar, Newman, Yuan, Yan, Zhang, Cosgrove, Manning,
  Ré, Acosta-Navas, Hudson, Zelikman, Durmus, Ladhak, Rong, Ren, Yao, Wang,
  Santhanam, Orr, Zheng, Yuksekgonul, Suzgun, Kim, Guha, Chatterji, Khattab,
  Henderson, Huang, Chi, Xie, Santurkar, Ganguli, Hashimoto, Icard, Zhang,
  Chaudhary, Wang, Li, Mai, Zhang, and Koreeda]{liang2023holistic}
Liang, P., Bommasani, R., Lee, T., Tsipras, D., Soylu, D., Yasunaga, M., Zhang,
  Y., Narayanan, D., Wu, Y., Kumar, A., Newman, B., Yuan, B., Yan, B., Zhang,
  C., Cosgrove, C., Manning, C.~D., Ré, C., Acosta-Navas, D., Hudson, D.~A.,
  Zelikman, E., Durmus, E., Ladhak, F., Rong, F., Ren, H., Yao, H., Wang, J.,
  Santhanam, K., Orr, L., Zheng, L., Yuksekgonul, M., Suzgun, M., Kim, N.,
  Guha, N., Chatterji, N., Khattab, O., Henderson, P., Huang, Q., Chi, R., Xie,
  S.~M., Santurkar, S., Ganguli, S., Hashimoto, T., Icard, T., Zhang, T.,
  Chaudhary, V., Wang, W., Li, X., Mai, Y., Zhang, Y., and Koreeda, Y.
\newblock Holistic evaluation of language models, 2023.

\bibitem[Liu et~al.(2021)Liu, Sap, Lu, Swayamdipta, Bhagavatula, Smith, and
  Choi]{Liu2021DExpertsDC}
Liu, A., Sap, M., Lu, X., Swayamdipta, S., Bhagavatula, C., Smith, N.~A., and
  Choi, Y.
\newblock Dexperts: Decoding-time controlled text generation with experts and
  anti-experts.
\newblock In \emph{Annual Meeting of the Association for Computational
  Linguistics}, 2021.
\newblock URL \url{https://api.semanticscholar.org/CorpusID:235313967}.

\bibitem[Liu et~al.(2022)Liu, Swayamdipta, Smith, and Choi]{liu2022wanli}
Liu, A., Swayamdipta, S., Smith, N.~A., and Choi, Y.
\newblock Wanli: Worker and ai collaboration for natural language inference
  dataset creation, 2022.

\bibitem[Liu et~al.(2024)Liu, Han, Wang, Tsvetkov, Choi, and
  Smith]{liu2024tuning}
Liu, A., Han, X., Wang, Y., Tsvetkov, Y., Choi, Y., and Smith, N.~A.
\newblock Tuning language models by proxy, 2024.

\bibitem[Liu et~al.(2023)Liu, Kumar, Liang, and Jia]{liu2023sampleefficient}
Liu, N.~F., Kumar, A., Liang, P., and Jia, R.
\newblock Are sample-efficient nlp models more robust?, 2023.

\bibitem[Long(2023)]{long2023large}
Long, J.
\newblock Large language model guided tree-of-thought.
\newblock \emph{arXiv preprint arXiv:2305.08291}, 2023.

\bibitem[Lu et~al.(2020)Lu, West, Zellers, Bras, Bhagavatula, and
  Choi]{Lu2020NeuroLogicD}
Lu, X., West, P., Zellers, R., Bras, R.~L., Bhagavatula, C., and Choi, Y.
\newblock Neurologic decoding: (un)supervised neural text generation with
  predicate logic constraints.
\newblock \emph{ArXiv}, abs/2010.12884, 2020.
\newblock URL \url{https://api.semanticscholar.org/CorpusID:225067055}.

\bibitem[Lu et~al.(2022)Lu, Welleck, Hessel, Jiang, Qin, West, Ammanabrolu, and
  Choi]{lu2022quark}
Lu, X., Welleck, S., Hessel, J., Jiang, L., Qin, L., West, P., Ammanabrolu, P.,
  and Choi, Y.
\newblock Quark: Controllable text generation with reinforced unlearning, 2022.

\bibitem[Ma et~al.(2023)Ma, Mishra, Liu, Su, Chen, Kulkarni, Cheng, Le, and
  Chi]{ma2023chatbots}
Ma, X., Mishra, S., Liu, A., Su, S., Chen, J., Kulkarni, C., Cheng, H.-T., Le,
  Q., and Chi, E.
\newblock Beyond chatbots: Explorellm for structured thoughts and personalized
  model responses, 2023.

\bibitem[MacAskill(2016)]{macaskill_normative_2016}
MacAskill, W.
\newblock Normative {Uncertainty} as a {Voting} {Problem}.
\newblock \emph{Mind}, 125\penalty0 (500):\penalty0 967--1004, October 2016.
\newblock ISSN 0026-4423.
\newblock \doi{10.1093/mind/fzv169}.
\newblock URL \url{https://doi.org/10.1093/mind/fzv169}.

\bibitem[Michael et~al.(2023)Michael, Mahdi, Rein, Petty, Dirani, Padmakumar,
  and Bowman]{michael2023debate}
Michael, J., Mahdi, S., Rein, D., Petty, J., Dirani, J., Padmakumar, V., and
  Bowman, S.~R.
\newblock Debate helps supervise unreliable experts.
\newblock \emph{arXiv preprint arXiv:2311.08702}, 2023.

\bibitem[Min et~al.(2020)Min, Michael, Hajishirzi, and
  Zettlemoyer]{min-etal-2020-ambigqa}
Min, S., Michael, J., Hajishirzi, H., and Zettlemoyer, L.
\newblock {A}mbig{QA}: Answering ambiguous open-domain questions.
\newblock In Webber, B., Cohn, T., He, Y., and Liu, Y. (eds.),
  \emph{Proceedings of the 2020 Conference on Empirical Methods in Natural
  Language Processing (EMNLP)}, pp.\  5783--5797, Online, November 2020.
  Association for Computational Linguistics.
\newblock \doi{10.18653/v1/2020.emnlp-main.466}.
\newblock URL \url{https://aclanthology.org/2020.emnlp-main.466}.

\bibitem[Mishra(2023)]{mishra2023ai}
Mishra, A.
\newblock Ai alignment and social choice: Fundamental limitations and policy
  implications, 2023.

\bibitem[Moulin(2004)]{moulin2004fair}
Moulin, H.
\newblock \emph{Fair Division and Collective Welfare}.
\newblock MIT Press, 2004.

\bibitem[Nagel(1979)]{nagel1979fragmentation}
Nagel, T.
\newblock The fragmentation of value.
\newblock In \emph{Mortal Questions}. Cambridge University Press, Cambridge,
  1979.

\bibitem[OpenAI(2023{\natexlab{a}})]{openai-davinci002}
OpenAI.
\newblock Openai davinci-002 model.
\newblock \url{https://www.openai.com}, 2023{\natexlab{a}}.
\newblock Accessed on Date 06/2023.

\bibitem[OpenAI(2023{\natexlab{b}})]{openai-gpt3.5-turbo}
OpenAI.
\newblock Openai gpt3.5-turbo.
\newblock \url{https://www.openai.com}, 2023{\natexlab{b}}.
\newblock Accessed on Date 06/2023.

\bibitem[Ouyang et~al.(2022)Ouyang, Wu, Jiang, Almeida, Wainwright, Mishkin,
  Zhang, Agarwal, Slama, Ray, Schulman, Hilton, Kelton, Miller, Simens, Askell,
  Welinder, Christiano, Leike, and Lowe]{ouyang2022training}
Ouyang, L., Wu, J., Jiang, X., Almeida, D., Wainwright, C.~L., Mishkin, P.,
  Zhang, C., Agarwal, S., Slama, K., Ray, A., Schulman, J., Hilton, J., Kelton,
  F., Miller, L., Simens, M., Askell, A., Welinder, P., Christiano, P., Leike,
  J., and Lowe, R.
\newblock Training language models to follow instructions with human feedback,
  2022.

\bibitem[Ovadya(2023)]{Ovadya2023}
Ovadya, A.
\newblock Reimagining democracy for ai.
\newblock \emph{Journal of Democracy}, 34\penalty0 (4):\penalty0 162--170, Oct
  2023.

\bibitem[Page(2008)]{page2008difference}
Page, S.
\newblock \emph{The difference: How the power of diversity creates better
  groups, firms, schools, and societies-new edition}.
\newblock Princeton University Press, 2008.

\bibitem[Page(2019)]{page2019diversity}
Page, S.~E.
\newblock \emph{The diversity bonus: How great teams pay off in the knowledge
  economy}.
\newblock Princeton University Press, 2019.

\bibitem[Pan et~al.(2023)Pan, Chan, Zou, Li, Basart, Woodside, Ng, Zhang,
  Emmons, and Hendrycks]{pan2023rewards}
Pan, A., Chan, J.~S., Zou, A., Li, N., Basart, S., Woodside, T., Ng, J., Zhang,
  H., Emmons, S., and Hendrycks, D.
\newblock Do the rewards justify the means? measuring trade-offs between
  rewards and ethical behavior in the machiavelli benchmark, 2023.

\bibitem[Park et~al.(2022)Park, Popowski, Cai, Morris, Liang, and
  Bernstein]{park2022social}
Park, J.~S., Popowski, L., Cai, C.~J., Morris, M.~R., Liang, P., and Bernstein,
  M.~S.
\newblock Social simulacra: Creating populated prototypes for social computing
  systems, 2022.

\bibitem[Park et~al.(2023)Park, O'Brien, Cai, Morris, Liang, and
  Bernstein]{park2023generative}
Park, J.~S., O'Brien, J., Cai, C.~J., Morris, M.~R., Liang, P., and Bernstein,
  M.~S.
\newblock Generative agents: Interactive simulacra of human behavior.
\newblock In \emph{Proceedings of the 36th Annual ACM Symposium on User
  Interface Software and Technology}, pp.\  1--22, 2023.

\bibitem[Peng et~al.(2023)Peng, Netanyahu, Ho, Shu, Bobu, Shah, and
  Agrawal]{peng2023diagnosis}
Peng, A., Netanyahu, A., Ho, M.~K., Shu, T., Bobu, A., Shah, J., and Agrawal,
  P.
\newblock Diagnosis, feedback, adaptation: A human-in-the-loop framework for
  test-time policy adaptation.
\newblock In \emph{Proceedings of the 40th International Conference on Machine
  Learning}, 2023.

\bibitem[Perez et~al.(2022)Perez, Ringer, Luko{\v{s}}i{\=u}t{\.e}, Nguyen,
  Chen, Heiner, Pettit, Olsson, Kundu, Kadavath, et~al.]{perez2022discovering}
Perez, E., Ringer, S., Luko{\v{s}}i{\=u}t{\.e}, K., Nguyen, K., Chen, E.,
  Heiner, S., Pettit, C., Olsson, C., Kundu, S., Kadavath, S., et~al.
\newblock Discovering language model behaviors with model-written evaluations.
\newblock \emph{arXiv preprint arXiv:2212.09251}, pp.\  13387--13434, July
  2022.
\newblock \doi{10.18653/v1/2023.findings-acl.847}.
\newblock URL \url{https://aclanthology.org/2023.findings-acl.847}.

\bibitem[{Pew Research Center}(2021)]{pewharassment2021}
{Pew Research Center}.
\newblock The state of online harassment.
\newblock Technical report, {Washington, D.C.}, January 2021.
\newblock URL
  \url{https://www.pewresearch.org/internet/2021/01/13/the-state-of-online-harassment/}.

\bibitem[Qin et~al.(2022)Qin, Welleck, Khashabi, and Choi]{Qin2022COLDDE}
Qin, L., Welleck, S., Khashabi, D., and Choi, Y.
\newblock Cold decoding: Energy-based constrained text generation with langevin
  dynamics.
\newblock \emph{ArXiv}, abs/2202.11705, 2022.
\newblock URL \url{https://api.semanticscholar.org/CorpusID:247058662}.

\bibitem[Ramezani \& Xu(2023)Ramezani and Xu]{ramezani2023knowledge}
Ramezani, A. and Xu, Y.
\newblock Knowledge of cultural moral norms in large language models, 2023.

\bibitem[Ramé et~al.(2023)Ramé, Couairon, Shukor, Dancette, Gaya, Soulier,
  and Cord]{ramé2023rewarded}
Ramé, A., Couairon, G., Shukor, M., Dancette, C., Gaya, J.-B., Soulier, L.,
  and Cord, M.
\newblock Rewarded soups: towards pareto-optimal alignment by interpolating
  weights fine-tuned on diverse rewards, 2023.

\bibitem[Ratcliffe(2016)]{AlbertEinstein}
Ratcliffe, S.
\newblock Albert einstein, 2016.
\newblock URL
  \url{https://www.oxfordreference.com/view/10.1093/acref/9780191826719.001.0001/q-oro-ed4-00003988}.

\bibitem[Rawls(1971)]{theoryofjustice}
Rawls, J.
\newblock \emph{A Theory of Justice: Original Edition}.
\newblock Harvard University Press, 1971.
\newblock ISBN 9780674880108.
\newblock URL \url{http://www.jstor.org/stable/j.ctvjf9z6v}.

\bibitem[Rawls(1996)]{rawls1996political}
Rawls, J.
\newblock \emph{Political Liberalism}.
\newblock Columbia University Press, New York, 1996.

\bibitem[Raz(1999)]{raz1999engaging}
Raz, J.
\newblock \emph{Engaging Reason: On the Theory of Value and Action}.
\newblock Oxford University Press, Oxford, 1999.

\bibitem[Santurkar et~al.(2023)Santurkar, Durmus, Ladhak, Lee, Liang, and
  Hashimoto]{santurkar2023opinions}
Santurkar, S., Durmus, E., Ladhak, F., Lee, C., Liang, P., and Hashimoto, T.
\newblock Whose opinions do language models reflect?, 2023.

\bibitem[Santy et~al.(2023)Santy, Liang, Bras, Reinecke, and
  Sap]{santy2023nlpositionality}
Santy, S., Liang, J.~T., Bras, R.~L., Reinecke, K., and Sap, M.
\newblock Nlpositionality: Characterizing design biases of datasets and models,
  2023.

\bibitem[Scherrer et~al.(2023)Scherrer, Shi, Feder, and
  Blei]{scherrer2023evaluating}
Scherrer, N., Shi, C., Feder, A., and Blei, D.~M.
\newblock Evaluating the moral beliefs encoded in llms, 2023.

\bibitem[Shajalal et~al.(2023)Shajalal, Atabuzzaman, Baby, Karim, and
  Boden]{Shajalal_2023}
Shajalal, M., Atabuzzaman, M., Baby, M.~B., Karim, M.~R., and Boden, A.
\newblock \emph{Textual Entailment Recognition with Semantic Features
  from Empirical Text Representation}, pp.\  183–195.
\newblock Springer International Publishing, 2023.
\newblock ISBN 9783031332319.
\newblock \doi{10.1007/978-3-031-33231-9_12}.
\newblock URL \url{http://dx.doi.org/10.1007/978-3-031-33231-9_12}.

\bibitem[Shanahan \& Clarke(2023)Shanahan and Clarke]{shanahan2023evaluating}
Shanahan, M. and Clarke, C.
\newblock Evaluating large language model creativity from a literary
  perspective, 2023.

\bibitem[Shanahan et~al.(2023)Shanahan, McDonell, and
  Reynolds]{shanahan2023roleplay}
Shanahan, M., McDonell, K., and Reynolds, L.
\newblock Role-play with large language models, 2023.

\bibitem[Sharma et~al.(2023{\natexlab{a}})Sharma, Lin, Miner, Atkins, and
  Althoff]{sharma-rephrase}
Sharma, A., Lin, I.~W., Miner, A.~S., Atkins, D.~C., and Althoff, T.
\newblock Human--ai collaboration enables more empathic conversations in
  text-based peer-to-peer mental health support.
\newblock \emph{Nature Machine Intelligence}, 5\penalty0 (1):\penalty0 46--57,
  2023{\natexlab{a}}.
\newblock \doi{10.1038/s42256-022-00593-2}.
\newblock URL \url{https://doi.org/10.1038/s42256-022-00593-2}.

\bibitem[Sharma et~al.(2023{\natexlab{b}})Sharma, Rushton, Lin, Wadden, Lucas,
  Miner, Nguyen, and Althoff]{sharma-etal-2023-cognitive}
Sharma, A., Rushton, K., Lin, I., Wadden, D., Lucas, K., Miner, A., Nguyen, T.,
  and Althoff, T.
\newblock Cognitive reframing of negative thoughts through human-language model
  interaction.
\newblock In Rogers, A., Boyd-Graber, J., and Okazaki, N. (eds.),
  \emph{Proceedings of the 61st Annual Meeting of the Association for
  Computational Linguistics (Volume 1: Long Papers)}, pp.\  9977--10000,
  Toronto, Canada, July 2023{\natexlab{b}}. Association for Computational
  Linguistics.
\newblock \doi{10.18653/v1/2023.acl-long.555}.
\newblock URL \url{https://aclanthology.org/2023.acl-long.555}.

\bibitem[Sharma et~al.(2023{\natexlab{c}})Sharma, Rushton, Lin, Nguyen, and
  Althoff]{Sharma2023FacilitatingSM}
Sharma, A., Rushton, K., Lin, I.~W., Nguyen, T., and Althoff, T.
\newblock Facilitating self-guided mental health interventions through
  human-language model interaction: A case study of cognitive restructuring.
\newblock \emph{ArXiv}, abs/2310.15461, 2023{\natexlab{c}}.
\newblock URL \url{https://api.semanticscholar.org/CorpusID:264439507}.

\bibitem[Sharma et~al.(2023{\natexlab{d}})Sharma, Rushton, Lin, Wadden, Lucas,
  Miner, Nguyen, and Althoff]{sharma2023cognitive}
Sharma, A., Rushton, K., Lin, I.~W., Wadden, D., Lucas, K.~G., Miner, A.~S.,
  Nguyen, T., and Althoff, T.
\newblock Cognitive reframing of negative thoughts through human-language model
  interaction, 2023{\natexlab{d}}.

\bibitem[Sher(1998)]{sher1998possibility}
Sher, G.
\newblock On the possibility of a substantive theory of truth.
\newblock \emph{Synthese}, 117:\penalty0 133–172, 1998.

\bibitem[Simmons(2023)]{simmons-2023-moral}
Simmons, G.
\newblock Moral mimicry: Large language models produce moral rationalizations
  tailored to political identity.
\newblock In Padmakumar, V., Vallejo, G., and Fu, Y. (eds.), \emph{Proceedings
  of the 61st Annual Meeting of the Association for Computational Linguistics
  (Volume 4: Student Research Workshop)}, pp.\  282--297, Toronto, Canada, July
  2023. Association for Computational Linguistics.
\newblock \doi{10.18653/v1/2023.acl-srw.40}.
\newblock URL \url{https://aclanthology.org/2023.acl-srw.40}.

\bibitem[Siththaranjan et~al.(2023)Siththaranjan, Laidlaw, and
  Hadfield-Menell]{siththaranjan2023distributional}
Siththaranjan, A., Laidlaw, C., and Hadfield-Menell, D.
\newblock Distributional preference learning: Understanding and accounting for
  hidden context in rlhf, 2023.

\bibitem[Solaiman \& Dennison(2021)Solaiman and Dennison]{solaiman2021process}
Solaiman, I. and Dennison, C.
\newblock Process for adapting language models to society (palms) with
  values-targeted datasets, 2021.

\bibitem[Song et~al.(2024)Song, Pendse, Kumar, and Choudhury]{song2024typing}
Song, I., Pendse, S.~R., Kumar, N., and Choudhury, M.~D.
\newblock The typing cure: Experiences with large language model chatbots for
  mental health support, 2024.

\bibitem[Sorensen et~al.(2023)Sorensen, Jiang, Hwang, Levine, Pyatkin, West,
  Dziri, Lu, Rao, Bhagavatula, Sap, Tasioulas, and Choi]{sorensen2023value}
Sorensen, T., Jiang, L., Hwang, J., Levine, S., Pyatkin, V., West, P., Dziri,
  N., Lu, X., Rao, K., Bhagavatula, C., Sap, M., Tasioulas, J., and Choi, Y.
\newblock Value kaleidoscope: Engaging ai with pluralistic human values,
  rights, and duties, 2023.

\bibitem[Srivastava et~al.(2023)Srivastava, Rastogi, Rao, Shoeb, Abid, Fisch,
  Brown, Santoro, Gupta, Garriga-Alonso, Kluska, Lewkowycz, Agarwal, Power,
  Ray, Warstadt, Kocurek, Safaya, Tazarv, Xiang, Parrish, Nie, Hussain, Askell,
  Dsouza, Slone, Rahane, Iyer, Andreassen, Madotto, Santilli, Stuhlmüller,
  Dai, La, Lampinen, Zou, Jiang, Chen, Vuong, Gupta, Gottardi, Norelli,
  Venkatesh, Gholamidavoodi, Tabassum, Menezes, Kirubarajan, Mullokandov,
  Sabharwal, Herrick, Efrat, Erdem, Karakaş, Roberts, Loe, Zoph, Bojanowski,
  Özyurt, Hedayatnia, Neyshabur, Inden, Stein, Ekmekci, Lin, Howald, Orinion,
  Diao, Dour, Stinson, Argueta, Ramírez, Singh, Rathkopf, Meng, Baral, Wu,
  Callison-Burch, Waites, Voigt, Manning, Potts, Ramirez, Rivera, Siro, Raffel,
  Ashcraft, Garbacea, Sileo, Garrette, Hendrycks, Kilman, Roth, Freeman,
  Khashabi, Levy, González, Perszyk, Hernandez, Chen, Ippolito, Gilboa, Dohan,
  Drakard, Jurgens, Datta, Ganguli, Emelin, Kleyko, Yuret, Chen, Tam, Hupkes,
  Misra, Buzan, Mollo, Yang, Lee, Schrader, Shutova, Cubuk, Segal, Hagerman,
  Barnes, Donoway, Pavlick, Rodola, Lam, Chu, Tang, Erdem, Chang, Chi, Dyer,
  Jerzak, Kim, Manyasi, Zheltonozhskii, Xia, Siar, Martínez-Plumed, Happé,
  Chollet, Rong, Mishra, Winata, de~Melo, Kruszewski, Parascandolo, Mariani,
  Wang, Jaimovitch-López, Betz, Gur-Ari, Galijasevic, Kim, Rashkin,
  Hajishirzi, Mehta, Bogar, Shevlin, Schütze, Yakura, Zhang, Wong, Ng, Noble,
  Jumelet, Geissinger, Kernion, Hilton, Lee, Fisac, Simon, Koppel, Zheng, Zou,
  Kocoń, Thompson, Wingfield, Kaplan, Radom, Sohl-Dickstein, Phang, Wei,
  Yosinski, Novikova, Bosscher, Marsh, Kim, Taal, Engel, Alabi, Xu, Song, Tang,
  Waweru, Burden, Miller, Balis, Batchelder, Berant, Frohberg, Rozen,
  Hernandez-Orallo, Boudeman, Guerr, Jones, Tenenbaum, Rule, Chua, Kanclerz,
  Livescu, Krauth, Gopalakrishnan, Ignatyeva, Markert, Dhole, Gimpel, Omondi,
  Mathewson, Chiafullo, Shkaruta, Shridhar, McDonell, Richardson, Reynolds,
  Gao, Zhang, Dugan, Qin, Contreras-Ochando, Morency, Moschella, Lam, Noble,
  Schmidt, He, Colón, Metz, Şenel, Bosma, Sap, ter Hoeve, Farooqi, Faruqui,
  Mazeika, Baturan, Marelli, Maru, Quintana, Tolkiehn, Giulianelli, Lewis,
  Potthast, Leavitt, Hagen, Schubert, Baitemirova, Arnaud, McElrath, Yee,
  Cohen, Gu, Ivanitskiy, Starritt, Strube, Swędrowski, Bevilacqua, Yasunaga,
  Kale, Cain, Xu, Suzgun, Walker, Tiwari, Bansal, Aminnaseri, Geva, Gheini, T,
  Peng, Chi, Lee, Krakover, Cameron, Roberts, Doiron, Martinez, Nangia,
  Deckers, Muennighoff, Keskar, Iyer, Constant, Fiedel, Wen, Zhang, Agha,
  Elbaghdadi, Levy, Evans, Casares, Doshi, Fung, Liang, Vicol,
  Alipoormolabashi, Liao, Liang, Chang, Eckersley, Htut, Hwang, Miłkowski,
  Patil, Pezeshkpour, Oli, Mei, Lyu, Chen, Banjade, Rudolph, Gabriel, Habacker,
  Risco, Millière, Garg, Barnes, Saurous, Arakawa, Raymaekers, Frank, Sikand,
  Novak, Sitelew, LeBras, Liu, Jacobs, Zhang, Salakhutdinov, Chi, Lee, Stovall,
  Teehan, Yang, Singh, Mohammad, Anand, Dillavou, Shleifer, Wiseman, Gruetter,
  Bowman, Schoenholz, Han, Kwatra, Rous, Ghazarian, Ghosh, Casey, Bischoff,
  Gehrmann, Schuster, Sadeghi, Hamdan, Zhou, Srivastava, Shi, Singh, Asaadi,
  Gu, Pachchigar, Toshniwal, Upadhyay, Shyamolima, Debnath, Shakeri, Thormeyer,
  Melzi, Reddy, Makini, Lee, Torene, Hatwar, Dehaene, Divic, Ermon, Biderman,
  Lin, Prasad, Piantadosi, Shieber, Misherghi, Kiritchenko, Mishra, Linzen,
  Schuster, Li, Yu, Ali, Hashimoto, Wu, Desbordes, Rothschild, Phan, Wang,
  Nkinyili, Schick, Kornev, Tunduny, Gerstenberg, Chang, Neeraj, Khot, Shultz,
  Shaham, Misra, Demberg, Nyamai, Raunak, Ramasesh, Prabhu, Padmakumar,
  Srikumar, Fedus, Saunders, Zhang, Vossen, Ren, Tong, Zhao, Wu, Shen,
  Yaghoobzadeh, Lakretz, Song, Bahri, Choi, Yang, Hao, Chen, Belinkov, Hou,
  Hou, Bai, Seid, Zhao, Wang, Wang, Wang, and Wu]{srivastava2023imitation}
Srivastava, A., Rastogi, A., Rao, A., Shoeb, A. A.~M., Abid, A., Fisch, A.,
  Brown, A.~R., Santoro, A., Gupta, A., Garriga-Alonso, A., Kluska, A.,
  Lewkowycz, A., Agarwal, A., Power, A., Ray, A., Warstadt, A., Kocurek, A.~W.,
  Safaya, A., Tazarv, A., Xiang, A., Parrish, A., Nie, A., Hussain, A., Askell,
  A., Dsouza, A., Slone, A., Rahane, A., Iyer, A.~S., Andreassen, A., Madotto,
  A., Santilli, A., Stuhlmüller, A., Dai, A., La, A., Lampinen, A., Zou, A.,
  Jiang, A., Chen, A., Vuong, A., Gupta, A., Gottardi, A., Norelli, A.,
  Venkatesh, A., Gholamidavoodi, A., Tabassum, A., Menezes, A., Kirubarajan,
  A., Mullokandov, A., Sabharwal, A., Herrick, A., Efrat, A., Erdem, A.,
  Karakaş, A., Roberts, B.~R., Loe, B.~S., Zoph, B., Bojanowski, B., Özyurt,
  B., Hedayatnia, B., Neyshabur, B., Inden, B., Stein, B., Ekmekci, B., Lin,
  B.~Y., Howald, B., Orinion, B., Diao, C., Dour, C., Stinson, C., Argueta, C.,
  Ramírez, C.~F., Singh, C., Rathkopf, C., Meng, C., Baral, C., Wu, C.,
  Callison-Burch, C., Waites, C., Voigt, C., Manning, C.~D., Potts, C.,
  Ramirez, C., Rivera, C.~E., Siro, C., Raffel, C., Ashcraft, C., Garbacea, C.,
  Sileo, D., Garrette, D., Hendrycks, D., Kilman, D., Roth, D., Freeman, D.,
  Khashabi, D., Levy, D., González, D.~M., Perszyk, D., Hernandez, D., Chen,
  D., Ippolito, D., Gilboa, D., Dohan, D., Drakard, D., Jurgens, D., Datta, D.,
  Ganguli, D., Emelin, D., Kleyko, D., Yuret, D., Chen, D., Tam, D., Hupkes,
  D., Misra, D., Buzan, D., Mollo, D.~C., Yang, D., Lee, D.-H., Schrader, D.,
  Shutova, E., Cubuk, E.~D., Segal, E., Hagerman, E., Barnes, E., Donoway, E.,
  Pavlick, E., Rodola, E., Lam, E., Chu, E., Tang, E., Erdem, E., Chang, E.,
  Chi, E.~A., Dyer, E., Jerzak, E., Kim, E., Manyasi, E.~E., Zheltonozhskii,
  E., Xia, F., Siar, F., Martínez-Plumed, F., Happé, F., Chollet, F., Rong,
  F., Mishra, G., Winata, G.~I., de~Melo, G., Kruszewski, G., Parascandolo, G.,
  Mariani, G., Wang, G., Jaimovitch-López, G., Betz, G., Gur-Ari, G.,
  Galijasevic, H., Kim, H., Rashkin, H., Hajishirzi, H., Mehta, H., Bogar, H.,
  Shevlin, H., Schütze, H., Yakura, H., Zhang, H., Wong, H.~M., Ng, I., Noble,
  I., Jumelet, J., Geissinger, J., Kernion, J., Hilton, J., Lee, J., Fisac,
  J.~F., Simon, J.~B., Koppel, J., Zheng, J., Zou, J., Kocoń, J., Thompson,
  J., Wingfield, J., Kaplan, J., Radom, J., Sohl-Dickstein, J., Phang, J., Wei,
  J., Yosinski, J., Novikova, J., Bosscher, J., Marsh, J., Kim, J., Taal, J.,
  Engel, J., Alabi, J., Xu, J., Song, J., Tang, J., Waweru, J., Burden, J.,
  Miller, J., Balis, J.~U., Batchelder, J., Berant, J., Frohberg, J., Rozen,
  J., Hernandez-Orallo, J., Boudeman, J., Guerr, J., Jones, J., Tenenbaum,
  J.~B., Rule, J.~S., Chua, J., Kanclerz, K., Livescu, K., Krauth, K.,
  Gopalakrishnan, K., Ignatyeva, K., Markert, K., Dhole, K.~D., Gimpel, K.,
  Omondi, K., Mathewson, K., Chiafullo, K., Shkaruta, K., Shridhar, K.,
  McDonell, K., Richardson, K., Reynolds, L., Gao, L., Zhang, L., Dugan, L.,
  Qin, L., Contreras-Ochando, L., Morency, L.-P., Moschella, L., Lam, L.,
  Noble, L., Schmidt, L., He, L., Colón, L.~O., Metz, L., Şenel, L.~K.,
  Bosma, M., Sap, M., ter Hoeve, M., Farooqi, M., Faruqui, M., Mazeika, M.,
  Baturan, M., Marelli, M., Maru, M., Quintana, M. J.~R., Tolkiehn, M.,
  Giulianelli, M., Lewis, M., Potthast, M., Leavitt, M.~L., Hagen, M.,
  Schubert, M., Baitemirova, M.~O., Arnaud, M., McElrath, M., Yee, M.~A.,
  Cohen, M., Gu, M., Ivanitskiy, M., Starritt, M., Strube, M., Swędrowski, M.,
  Bevilacqua, M., Yasunaga, M., Kale, M., Cain, M., Xu, M., Suzgun, M., Walker,
  M., Tiwari, M., Bansal, M., Aminnaseri, M., Geva, M., Gheini, M., T, M.~V.,
  Peng, N., Chi, N.~A., Lee, N., Krakover, N. G.-A., Cameron, N., Roberts, N.,
  Doiron, N., Martinez, N., Nangia, N., Deckers, N., Muennighoff, N., Keskar,
  N.~S., Iyer, N.~S., Constant, N., Fiedel, N., Wen, N., Zhang, O., Agha, O.,
  Elbaghdadi, O., Levy, O., Evans, O., Casares, P. A.~M., Doshi, P., Fung, P.,
  Liang, P.~P., Vicol, P., Alipoormolabashi, P., Liao, P., Liang, P., Chang,
  P., Eckersley, P., Htut, P.~M., Hwang, P., Miłkowski, P., Patil, P.,
  Pezeshkpour, P., Oli, P., Mei, Q., Lyu, Q., Chen, Q., Banjade, R., Rudolph,
  R.~E., Gabriel, R., Habacker, R., Risco, R., Millière, R., Garg, R., Barnes,
  R., Saurous, R.~A., Arakawa, R., Raymaekers, R., Frank, R., Sikand, R.,
  Novak, R., Sitelew, R., LeBras, R., Liu, R., Jacobs, R., Zhang, R.,
  Salakhutdinov, R., Chi, R., Lee, R., Stovall, R., Teehan, R., Yang, R.,
  Singh, S., Mohammad, S.~M., Anand, S., Dillavou, S., Shleifer, S., Wiseman,
  S., Gruetter, S., Bowman, S.~R., Schoenholz, S.~S., Han, S., Kwatra, S.,
  Rous, S.~A., Ghazarian, S., Ghosh, S., Casey, S., Bischoff, S., Gehrmann, S.,
  Schuster, S., Sadeghi, S., Hamdan, S., Zhou, S., Srivastava, S., Shi, S.,
  Singh, S., Asaadi, S., Gu, S.~S., Pachchigar, S., Toshniwal, S., Upadhyay,
  S., Shyamolima, Debnath, Shakeri, S., Thormeyer, S., Melzi, S., Reddy, S.,
  Makini, S.~P., Lee, S.-H., Torene, S., Hatwar, S., Dehaene, S., Divic, S.,
  Ermon, S., Biderman, S., Lin, S., Prasad, S., Piantadosi, S.~T., Shieber,
  S.~M., Misherghi, S., Kiritchenko, S., Mishra, S., Linzen, T., Schuster, T.,
  Li, T., Yu, T., Ali, T., Hashimoto, T., Wu, T.-L., Desbordes, T., Rothschild,
  T., Phan, T., Wang, T., Nkinyili, T., Schick, T., Kornev, T., Tunduny, T.,
  Gerstenberg, T., Chang, T., Neeraj, T., Khot, T., Shultz, T., Shaham, U.,
  Misra, V., Demberg, V., Nyamai, V., Raunak, V., Ramasesh, V., Prabhu, V.~U.,
  Padmakumar, V., Srikumar, V., Fedus, W., Saunders, W., Zhang, W., Vossen, W.,
  Ren, X., Tong, X., Zhao, X., Wu, X., Shen, X., Yaghoobzadeh, Y., Lakretz, Y.,
  Song, Y., Bahri, Y., Choi, Y., Yang, Y., Hao, Y., Chen, Y., Belinkov, Y.,
  Hou, Y., Hou, Y., Bai, Y., Seid, Z., Zhao, Z., Wang, Z., Wang, Z.~J., Wang,
  Z., and Wu, Z.
\newblock Beyond the imitation game: Quantifying and extrapolating the
  capabilities of language models, 2023.

\bibitem[Sumers et~al.(2023)Sumers, Yao, Narasimhan, and
  Griffiths]{sumers2023cognitive}
Sumers, T.~R., Yao, S., Narasimhan, K., and Griffiths, T.~L.
\newblock Cognitive architectures for language agents, 2023.

\bibitem[Swamy et~al.(2024)Swamy, Dann, Kidambi, Wu, and
  Agarwal]{swamy2024minimaximalist}
Swamy, G., Dann, C., Kidambi, R., Wu, Z.~S., and Agarwal, A.
\newblock A minimaximalist approach to reinforcement learning from human
  feedback, 2024.

\bibitem[Taori et~al.(2023)Taori, Gulrajani, Zhang, Dubois, Li, Guestrin,
  Liang, and Hashimoto]{alpaca}
Taori, R., Gulrajani, I., Zhang, T., Dubois, Y., Li, X., Guestrin, C., Liang,
  P., and Hashimoto, T.~B.
\newblock Stanford alpaca: An instruction-following llama model.
\newblock \url{https://github.com/tatsu-lab/stanford_alpaca}, 2023.

\bibitem[Tasioulas(2022)]{tasioulas}
Tasioulas, J.
\newblock {Artificial Intelligence, Humanistic Ethics}.
\newblock \emph{Daedalus}, 151\penalty0 (2):\penalty0 232--243, 05 2022.
\newblock ISSN 0011-5266.
\newblock \doi{10.1162/daed_a_01912}.
\newblock URL \url{https://doi.org/10.1162/daed\_a\_01912}.

\bibitem[Team et~al.(2024)Team, Mesnard, Hardin, Dadashi, Bhupatiraju, Pathak,
  Sifre, Rivière, Kale, Love, Tafti, Hussenot, Chowdhery, Roberts, Barua,
  Botev, Castro-Ros, Slone, Héliou, Tacchetti, Bulanova, Paterson, Tsai,
  Shahriari, Lan, Choquette-Choo, Crepy, Cer, Ippolito, Reid, Buchatskaya, Ni,
  Noland, Yan, Tucker, Muraru, Rozhdestvenskiy, Michalewski, Tenney,
  Grishchenko, Austin, Keeling, Labanowski, Lespiau, Stanway, Brennan, Chen,
  Ferret, Chiu, Mao-Jones, Lee, Yu, Millican, Sjoesund, Lee, Dixon, Reid,
  Mikuła, Wirth, Sharman, Chinaev, Thain, Bachem, Chang, Wahltinez, Bailey,
  Michel, Yotov, Sessa, Chaabouni, Comanescu, Jana, Anil, McIlroy, Liu,
  Mullins, Smith, Borgeaud, Girgin, Douglas, Pandya, Shakeri, De, Klimenko,
  Hennigan, Feinberg, Stokowiec, hui Chen, Ahmed, Gong, Warkentin, Peran,
  Giang, Farabet, Vinyals, Dean, Kavukcuoglu, Hassabis, Ghahramani, Eck,
  Barral, Pereira, Collins, Joulin, Fiedel, Senter, Andreev, and
  Kenealy]{gemmateam2024gemma}
Team, G., Mesnard, T., Hardin, C., Dadashi, R., Bhupatiraju, S., Pathak, S.,
  Sifre, L., Rivière, M., Kale, M.~S., Love, J., Tafti, P., Hussenot, L.,
  Chowdhery, A., Roberts, A., Barua, A., Botev, A., Castro-Ros, A., Slone, A.,
  Héliou, A., Tacchetti, A., Bulanova, A., Paterson, A., Tsai, B., Shahriari,
  B., Lan, C.~L., Choquette-Choo, C.~A., Crepy, C., Cer, D., Ippolito, D.,
  Reid, D., Buchatskaya, E., Ni, E., Noland, E., Yan, G., Tucker, G., Muraru,
  G.-C., Rozhdestvenskiy, G., Michalewski, H., Tenney, I., Grishchenko, I.,
  Austin, J., Keeling, J., Labanowski, J., Lespiau, J.-B., Stanway, J.,
  Brennan, J., Chen, J., Ferret, J., Chiu, J., Mao-Jones, J., Lee, K., Yu, K.,
  Millican, K., Sjoesund, L.~L., Lee, L., Dixon, L., Reid, M., Mikuła, M.,
  Wirth, M., Sharman, M., Chinaev, N., Thain, N., Bachem, O., Chang, O.,
  Wahltinez, O., Bailey, P., Michel, P., Yotov, P., Sessa, P.~G., Chaabouni,
  R., Comanescu, R., Jana, R., Anil, R., McIlroy, R., Liu, R., Mullins, R.,
  Smith, S.~L., Borgeaud, S., Girgin, S., Douglas, S., Pandya, S., Shakeri, S.,
  De, S., Klimenko, T., Hennigan, T., Feinberg, V., Stokowiec, W., hui Chen,
  Y., Ahmed, Z., Gong, Z., Warkentin, T., Peran, L., Giang, M., Farabet, C.,
  Vinyals, O., Dean, J., Kavukcuoglu, K., Hassabis, D., Ghahramani, Z., Eck,
  D., Barral, J., Pereira, F., Collins, E., Joulin, A., Fiedel, N., Senter, E.,
  Andreev, A., and Kenealy, K.
\newblock Gemma: Open models based on gemini research and technology, 2024.

\bibitem[T{\"o}rnberg et~al.(2023)T{\"o}rnberg, Valeeva, Uitermark, and
  Bail]{tornberg2023simulating}
T{\"o}rnberg, P., Valeeva, D., Uitermark, J., and Bail, C.
\newblock Simulating social media using large language models to evaluate
  alternative news feed algorithms.
\newblock \emph{arXiv preprint arXiv:2310.05984}, 2023.

\bibitem[Touvron et~al.(2023)Touvron, Martin, Stone, Albert, Almahairi, Babaei,
  Bashlykov, Batra, Bhargava, Bhosale, Bikel, Blecher, Ferrer, Chen, Cucurull,
  Esiobu, Fernandes, Fu, Fu, Fuller, Gao, Goswami, Goyal, Hartshorn, Hosseini,
  Hou, Inan, Kardas, Kerkez, Khabsa, Kloumann, Korenev, Koura, Lachaux, Lavril,
  Lee, Liskovich, Lu, Mao, Martinet, Mihaylov, Mishra, Molybog, Nie, Poulton,
  Reizenstein, Rungta, Saladi, Schelten, Silva, Smith, Subramanian, Tan, Tang,
  Taylor, Williams, Kuan, Xu, Yan, Zarov, Zhang, Fan, Kambadur, Narang,
  Rodriguez, Stojnic, Edunov, and Scialom]{touvron2023llama}
Touvron, H., Martin, L., Stone, K., Albert, P., Almahairi, A., Babaei, Y.,
  Bashlykov, N., Batra, S., Bhargava, P., Bhosale, S., Bikel, D., Blecher, L.,
  Ferrer, C.~C., Chen, M., Cucurull, G., Esiobu, D., Fernandes, J., Fu, J., Fu,
  W., Fuller, B., Gao, C., Goswami, V., Goyal, N., Hartshorn, A., Hosseini, S.,
  Hou, R., Inan, H., Kardas, M., Kerkez, V., Khabsa, M., Kloumann, I., Korenev,
  A., Koura, P.~S., Lachaux, M.-A., Lavril, T., Lee, J., Liskovich, D., Lu, Y.,
  Mao, Y., Martinet, X., Mihaylov, T., Mishra, P., Molybog, I., Nie, Y.,
  Poulton, A., Reizenstein, J., Rungta, R., Saladi, K., Schelten, A., Silva,
  R., Smith, E.~M., Subramanian, R., Tan, X.~E., Tang, B., Taylor, R.,
  Williams, A., Kuan, J.~X., Xu, P., Yan, Z., Zarov, I., Zhang, Y., Fan, A.,
  Kambadur, M., Narang, S., Rodriguez, A., Stojnic, R., Edunov, S., and
  Scialom, T.
\newblock Llama 2: Open foundation and fine-tuned chat models, 2023.
\newblock URL \url{https://api.semanticscholar.org/CorpusID:259950998}.

\bibitem[Tozer et~al.(2017)Tozer, Mazzuchi, and Sarkani]{TOZER2017371}
Tozer, B., Mazzuchi, T., and Sarkani, S.
\newblock Many-objective stochastic path finding using reinforcement learning.
\newblock \emph{Expert Systems with Applications}, 72:\penalty0 371--382, 2017.
\newblock ISSN 0957-4174.
\newblock \doi{https://doi.org/10.1016/j.eswa.2016.10.045}.
\newblock URL
  \url{https://www.sciencedirect.com/science/article/pii/S0957417416305863}.

\bibitem[Wang et~al.(2023{\natexlab{a}})Wang, Ivison, Dasigi, Hessel, Khot,
  Chandu, Wadden, MacMillan, Smith, Beltagy, and Hajishirzi]{Wang2023HowFC}
Wang, Y., Ivison, H., Dasigi, P., Hessel, J., Khot, T., Chandu, K.~R., Wadden,
  D., MacMillan, K., Smith, N.~A., Beltagy, I., and Hajishirzi, H.
\newblock How far can camels go? exploring the state of instruction tuning on
  open resources.
\newblock \emph{ArXiv}, abs/2306.04751, 2023{\natexlab{a}}.
\newblock URL \url{https://api.semanticscholar.org/CorpusID:259108263}.

\bibitem[Wang et~al.(2023{\natexlab{b}})Wang, Zhong, Li, Mi, Zeng, Huang,
  Shang, Jiang, and Liu]{wang2023aligning}
Wang, Y., Zhong, W., Li, L., Mi, F., Zeng, X., Huang, W., Shang, L., Jiang, X.,
  and Liu, Q.
\newblock Aligning large language models with human: A survey,
  2023{\natexlab{b}}.

\bibitem[Wojcik et~al.(2022)Wojcik, Hilgard, Judd, Mocanu, Ragain, Hunzaker,
  Coleman, and Baxter]{wojcik2022birdwatch}
Wojcik, S., Hilgard, S., Judd, N., Mocanu, D., Ragain, S., Hunzaker, M. B.~F.,
  Coleman, K., and Baxter, J.
\newblock Birdwatch: Crowd wisdom and bridging algorithms can inform
  understanding and reduce the spread of misinformation, 2022.

\bibitem[Wortsman et~al.(2022)Wortsman, Ilharco, Gadre, Roelofs, Gontijo-Lopes,
  Morcos, Namkoong, Farhadi, Carmon, Kornblith, and Schmidt]{wortsman2022model}
Wortsman, M., Ilharco, G., Gadre, S.~Y., Roelofs, R., Gontijo-Lopes, R.,
  Morcos, A.~S., Namkoong, H., Farhadi, A., Carmon, Y., Kornblith, S., and
  Schmidt, L.
\newblock Model soups: averaging weights of multiple fine-tuned models improves
  accuracy without increasing inference time, 2022.

\bibitem[Wright(1992)]{wright1992truth}
Wright, C.
\newblock \emph{Truth and Objectivity}.
\newblock Harvard University Press, Cambridge, MA, 1992.

\bibitem[Yang et~al.(2019)Yang, Sun, and Narasimhan]{yang2019generalized}
Yang, R., Sun, X., and Narasimhan, K.
\newblock A generalized algorithm for multi-objective reinforcement learning
  and policy adaptation, 2019.

\bibitem[Zhao et~al.(2023)Zhao, Dang, and Grover]{zhao_group_2023}
Zhao, S., Dang, J., and Grover, A.
\newblock Group {Preference} {Optimization}: {Few}-{Shot} {Alignment} of
  {Large} {Language} {Models}.
\newblock 2023.
\newblock \doi{10.48550/ARXIV.2310.11523}.
\newblock URL \url{https://arxiv.org/abs/2310.11523}.
\newblock Publisher: arXiv Version Number: 1.

\bibitem[Zhou et~al.(2024)Zhou, Hwang, Ren, and Sap]{zhou2024relying}
Zhou, K., Hwang, J.~D., Ren, X., and Sap, M.
\newblock Relying on the unreliable: The impact of language models' reluctance
  to express uncertainty, 2024.

\bibitem[Ziems et~al.(2023)Ziems, Held, Shaikh, Chen, Zhang, and
  Yang]{ziems2023can}
Ziems, C., Held, W., Shaikh, O., Chen, J., Zhang, Z., and Yang, D.
\newblock Can large language models transform computational social science?
\newblock \emph{arXiv preprint arXiv:2305.03514}, 2023.

\bibitem[Zintgraf et~al.(2015)Zintgraf, Kanters, Roijers, Oliehoek, and
  Beau]{Zintgraf2015QualityAO}
Zintgraf, L.~M., Kanters, T.~V., Roijers, D.~M., Oliehoek, F.~A., and Beau, P.
\newblock Quality assessment of morl algorithms: A utility-based approach.
\newblock 2015.
\newblock URL \url{https://api.semanticscholar.org/CorpusID:15373186}.

\end{thebibliography}
